\documentclass[10pt,twocolumn,letterpaper]{article}

\usepackage[pagenumbers]{cvpr} 
\usepackage{booktabs}
\usepackage{graphicx}
\usepackage[table]{xcolor}

\usepackage{multirow}

\usepackage{array}
\usepackage{tabularx}

\usepackage{pifont}
\usepackage[table]{xcolor}

\definecolor{groupgray}{RGB}{240,240,240}
\definecolor{gainred}{RGB}{170,45,45}

\definecolor{refblue}{RGB}{0,120,255}

\newcommand{\cmark}{\ding{51}}
\newcommand{\xmark}{\ding{55}}

\definecolor{cvprblue}{rgb}{0.21,0.49,0.74}
\usepackage[pagebackref,breaklinks,colorlinks,allcolors=cvprblue]{hyperref}

\def\confName{CVPR}
\def\confYear{2027}

\title{SVA-WAM: Learning Future Representations for World Action Models \\via Subgoals and Visual-Action}

\author{Wenqi Zhu$^{1}$\thanks{This work was done during the internship at Dexmal.} ,
Fan Jia$^{2}$, Tiancai Wang$^{2}$, Jiale Cao$^{1}$\thanks{Corresponding author: Jiale Cao}\\
$^1$Tianjin University~~~$^2$Dexmal\\
{\tt\small \{wenqizhu,connor\}@tju.edu.cn, } 
{\tt\small \{jiafan,wtc\}@dexmal.com} 
}

\begin{document}
\maketitle
\begin{abstract}

World Action Models (WAMs) predict future visual dynamics and robot actions, using future representations to guide action generation. Effective guidance requires richer task and robot motion information in these future representations. However, most existing methods predict only short-horizon futures, providing limited information about later task stages. Moreover, future visual dynamics capture robot motion while including background and distractor objects that may interfere with action generation. To address these issues, we propose \textbf{SVA-WAM}, a framework for learning future representations for WAMs via long-horizon subgoal learning (LSL) and visual-action bridging (VAB). LSL captures long-horizon task progression, enriching future representations with information about later task stages. VAB introduces a robot-centric visual-action representation and inserts a visual-action expert between the video and action experts to connect predicted visual dynamics to robot actions. Together, these components guide action generation with task-aware and robot-centric future representations. Extensive experiments on both simulation benchmarks and real-world environments show the effectiveness of our method. For instance, SVA-WAM achieves a success rate of 74.8\% on LIBERO-Plus, outperforming the baseline by 21.0\%, highlighting strong generalization under environmental perturbations. On the long-horizon benchmark RMBench, SVA-WAM outperforms the baseline by 11.1\%, demonstrating its ability to complete long-horizon tasks. Furthermore, real-world experiments demonstrate reliable long-horizon manipulation and robustness to environmental variations. The project is available \href{https://zwq456.github.io/SVA-WAM_web}{here}.

\end{abstract}    
\section{Introduction}
\label{sec:intro}

\begin{figure}[t]
\centering
\includegraphics[width=1.0\linewidth]{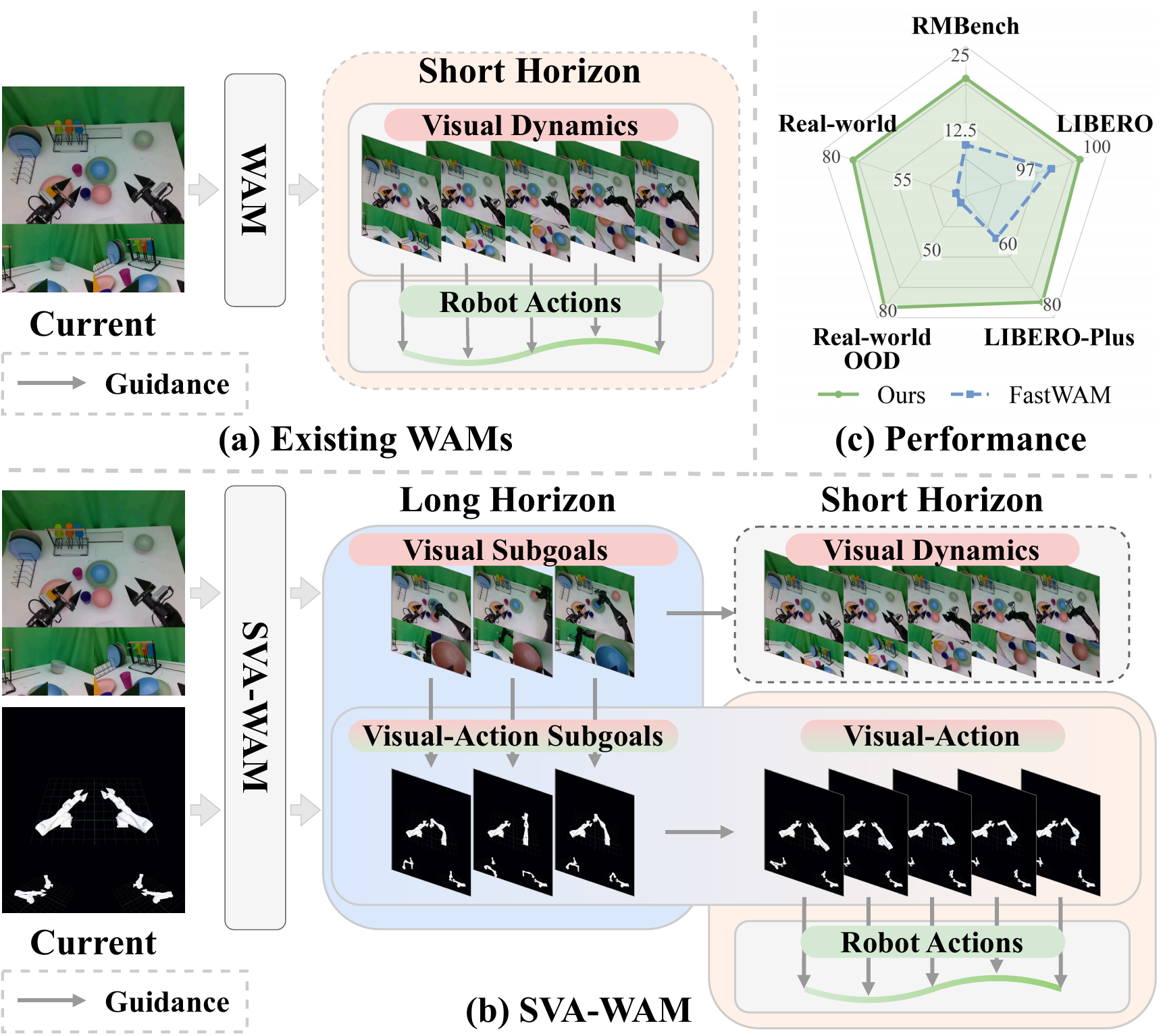}
\caption{\textbf{Comparison with existing WAMs.}
In (a), Existing WAMs jointly model short-horizon dynamics and actions, offering limited long-horizon task foresight while facing content interference and a representation gap between visual dynamics and actions.
In (b), SVA-WAM uses long-horizon subgoal learning to guide dense short-horizon prediction and introduces visual-action bridging based on multi-view visual-action representations to bridge visual dynamics and robot actions.
In (c), our SVA-WAM outperforms the baseline Fast-WAM \cite{yuan2026fastwam} across simulation benchmarks and real-world evaluations, demonstrating improved long-horizon manipulation and generalization.}

\label{fig:intro}
\end{figure}

 In Embodied AI, Vision Language Action models (VLAs) and World Action Models (WAMs) have emerged as two prominent approaches. VLAs~\cite{sun2025geovla,shi2026memoryvla,guo2026priorvla} typically generate robot actions directly from visual observations and language instructions, while WAMs~\cite{bi2025motus,li2026lingbotva,ye2026dreamzero} jointly model future visual dynamics and robot actions. Compared to VLAs, WAMs provide more fine-grained information about future states to guide action generation. WAMs can be mainly categorized into cascaded WAMs and joint WAMs~\cite{wang2605world}. Cascaded WAMs first predict future visual dynamics and then generate robot actions from these predictions. For example, VPP~\cite{hu2025vpp} and mimic-video~\cite{pai2025mimic} generate actions from predicted future latents and intermediate video features, respectively. In contrast, joint WAMs jointly model future visual dynamics and robot actions within a unified framework. For instance, DreamZero~\cite{ye2026dreamzero} and Motus~\cite{bi2025motus} adopt unified and multi stream architectures, respectively. However, whether future prediction can improve robot control depends on what information the predicted representations encode and how effectively that information guides action generation. We argue that effective future representations should capture both task-level and robot motion information. The task-level information specifies the target for task completion and the robot motion information provides direct guidance for action generation.

However, as shown in Fig.~\ref{fig:intro}(a), most existing WAMs~\cite{bi2025motus,li2026lingbotva,ye2026dreamzero} rely on dense short-horizon prediction. These predictions primarily capture local state changes, making it difficult to model later task stages and connect current actions to long-horizon task progress. Moreover, most existing WAMs predict future visual dynamics or incorporate additional information such as semantics~\cite{wang2026stwam} and depth~\cite{ma2026geosemwam}. These representations capture robot motion while also including scene details, such as backgrounds and distractor objects, that may interfere with action generation.

To address these limitations, as shown in Fig.~\ref{fig:intro}(b), we propose \textbf{SVA-WAM}, a framework for learning future representations for WAMs via long-horizon subgoal learning (LSL) and visual-action bridging (VAB). LSL learns to represent later task stages as long-horizon subgoals. These subgoals provide information about how the task is expected to progress and guide short-horizon predictions of visual dynamics and robot actions. Inspired by GeniWorld~\cite{gu2026geniworld}, we design a robot-centric visual-action representation that only encodes robot motion. VAB introduces a visual-action expert between the video and action experts to connect predicted future representations with robot actions. This design makes motion information more directly available to the action expert and helps reduce the influence of task-irrelevant scene details on action generation. We evaluate the performance of SVA-WAM on multiple simulation benchmarks, including LIBERO~\cite{liu2023libero}, LIBERO-Plus~\cite{fei2026liberoplus}, and RMBench~\cite{chen2026rmbench}, and real-world robot tasks. Experimental results significantly demonstrate the effectiveness and robustness of our method. Our main contributions can be summarized as follows:
\begin{itemize}

    \item We propose a joint long-horizon subgoal and visual-action guidance framework, which can improve  long-horizon task execution and generalization ability.
    
    \item We propose long-horizon subgoal learning to guide  short-horizon dynamics and actions prediction. We further propose visual-action bridging to better transfer task-relevant information from visual dynamics to robot actions using a robot-centric visual-action representation.
    
    \item Our SVA-WAM achieves superior performance across multiple simulation benchmarks and real-world tasks. For instance, SVA-WAM achieves 98.8\% and 74.8\% success rates on LIBERO and LIBERO-Plus, exceeding the baseline by 1.2\% and 21.0\%, respectively. Further, on long-horizon RMBench, it improves the baseline by 11.1\%. Real-world experiments further demonstrate reliable long-horizon manipulation and generalization.

\end{itemize}

\section{Related Work}
\label{sec:related}

\subsection{World Action Models}

World Action Models (WAMs)~\cite{bi2025motus,li2026lingbotva,ye2026dreamzero} combine future visual dynamics prediction with robot action generation to guide robot control. They can be mainly categorized into cascaded WAMs and joint WAMs~\cite{wang2605world}. Cascaded WAMs decouple visual dynamics modeling from action generation, using explicit future visual dynamics or learned future latent representations to decode action. UniPi~\cite{du2023unipi} generates text-conditioned video plans and extracts executable actions through an inverse dynamics model (IDM). LAPA~\cite{ye2025lapa} learns discrete latent actions from unlabeled videos, pretrains a vision-language-action model to predict these latent actions, and then fine-tunes it on a small amount of robot action data to generate executable actions. In contrast, joint WAMs model future visual dynamics and robot actions within a unified framework. Motus~\cite{bi2025motus} uses a three-expert Mixture-of-Transformers (MoT)~\cite{liang2025mixtureoftransformers} architecture to model semantic information, visual dynamics and actions. Cosmos Policy~\cite{kim2026cosmospolicy} adds value prediction to support control, world modeling, and planning. LingBot-VA~\cite{li2026lingbotva} unifies video prediction and action generation within a causal sequence through autoregressive diffusion. DreamZero~\cite{ye2026dreamzero} explores data efficiency and generalization through joint video and action modeling. Fast-WAM~\cite{yuan2026fastwam} improves inference efficiency by avoiding future video generation at test time. Recent studies also incorporate semantic features~\cite{wang2026stwam}, depth and geometry features~\cite{ma2026geosemwam,zhao2026sgwam,zhao2026rynnworld4d}, object and robot masks~\cite{yu2026maskwam}, or optical-flow features~\cite{chen2026flowwam} into WAMs to better guide action generation.

\subsection{Visual Planning}

Long-horizon prediction captures task progression and guides action generation toward task goals. World models can provide this guidance by predicting visual dynamics over longer time spans or learning sparse subgoals~\cite{fu2026novaplan,long2026scaling}. Visual subgoals serve as explicit future-state targets that connect long-horizon planning with action generation. In VLAs, ForeAct~\cite{zhang2026foreact} and Anticipation-VLA~\cite{zhang2026anticipationvla} plan subtasks and predict visual subgoals to guide local actions. StructVLA~\cite{jin2026structvla} predicts sparse visual milestones derived from robot kinematics. These methods complement current observations with anticipated task states. Recent WAMs also incorporate subgoal modeling. WLA~\cite{yang2026wla} jointly predicts textual subtasks, visual subgoals, and actions. ImageWAM~\cite{zhang2026imagewam} predicts future endpoints through image editing and uses the resulting features to guide actions. StageWAM~\cite{liu2026stagewam} predicts latent visual features of the next-stage subgoal. ProWAM~\cite{zhang2026prowam} jointly predicts actions and ordered sparse visual subgoals to provide progress-aware guidance.

SVA-WAM selects the terminal  of each subtask as a subgoal and predicts a sparse sequence of subgoals. WLA~\cite{yang2026wla} and ImageWAM~\cite{zhang2026imagewam} use a single subgoal at a fixed temporal interval from the current observation, and ProWAM~\cite{zhang2026prowam} uniformly samples future frames over a longer horizon. Our subgoal sequence captures task progression across multiple stages over a long horizon.

\subsection{Bridging Vision and Action}

Visual dynamics describe scene changes in image space, while robot actions are usually numerical vectors. This representation gap makes it difficult to translate visual predictions into actions. Moreover, task-irrelevant content, such as backgrounds and unrelated objects, may interfere with action generation. Recent WAMs explore alternative representations to connect future prediction with action generation. FlowWAM~\cite{chen2026flowwam} uses optical-flow videos for action prediction and action-conditioned world generation, while RepWAM~\cite{wang2026repwam} aligns visual and latent action tokens in a shared semantic space. Action Images~\cite{zhen2026actionimages} maps translation, rotation, and gripper states into multi-view images for world-model prediction and subsequent decoding into numerical actions. In action-conditioned world models, EA-WM~\cite{yang2026eawm} projects robot kinematics and actions into structured visual fields. GeniWorld~\cite{gu2026geniworld} takes numerical actions from an external policy and uses URDF geometry to render the corresponding robot motion as camera-aligned visual actions. The world model concatenates the latents of these visual actions with video latents along the channel dimension to predict how the scene will change when those actions are executed.

In SVA-WAM, we render robot motion from a separate viewpoint, predict future visual-action, and use these predictions as an intermediate modality that bridges visual observations and action generation.

\begin{figure*}[t]
\centering
\includegraphics[width=0.95\linewidth]{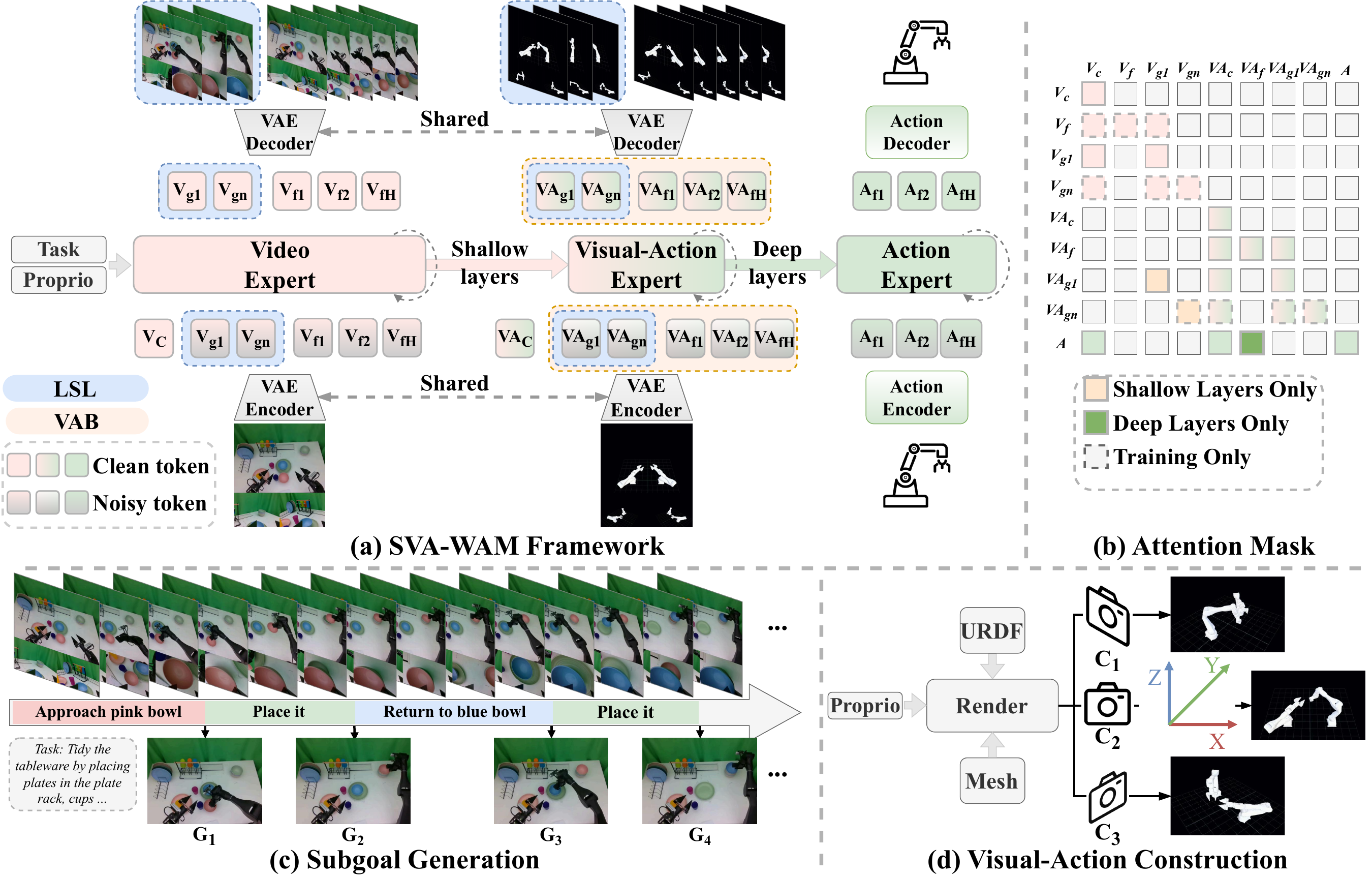}
\caption{\textbf{Overview of SVA-WAM.}
\textbf{(a) SVA-WAM framework.} A MoT-style network jointly models future dynamics, visual-action, and robot actions through three experts. The video and visual-action experts use a shared frozen VAE and predict both short-horizon dynamics and long-horizon subgoals. The visual-action expert bridges visual dynamics and action generation. 
\textbf{(b) Attention mask.} Information exchange is controlled by a structured layer-grouped mask. Long-horizon subgoals guide short-horizon prediction. Visual-action tokens attend to video tokens only in shallow layers, while action tokens attend to visual-action tokens only in deep layers.
\textbf{(c) Subgoal Generation.} Each demonstration trajectory is divided into segments according to task stages, and the final frame of each segment is annotated as a visual subgoal (\(G_1 \dots G_k\)) to capture long-horizon task progression.
\textbf{(d) Visual-action construction.} Given the current joint state \(q_c\), the robot meshes are rendered from three fixed orthogonal viewpoints using the URDF-defined kinematic structure to construct visual-action representations.}

\label{fig:method}
\end{figure*}

\section{Method}

\subsection{Overview}
\noindent\textbf{Preliminaries.}
WAMs jointly predict future visual dynamics and robot actions. Given a current visual observation \(O_c\), robot state $q_c$ and language instruction \(l\), WAMs first encode them into latent representations using a VAE encoder \(\mathcal E_{\mathrm V}\), action encoder \(\mathcal E_{\mathrm A}\) and a text encoder \(\mathcal E_{\mathrm T}\), respectively:
\begin{equation}
    \mathbf V_c = \mathcal E_{\mathrm V}(O_c),
    \qquad
    \mathbf Q_c = \mathcal E_{\mathrm A}(q_c),
    \qquad
    \mathbf T_l = \mathcal E_{\mathrm T}(l).
    \label{eq:condition_encoding}
\end{equation}

During training, \(O_c\) and ground-truth future observations \(\mathbf O^\star_{f_{1:H}}\) are jointly encoded by \(\mathcal E_{\mathrm V}\), while ground-truth actions \(\mathbf a^\star_{f_{1:H}}\) are encoded by \(\mathcal E_{\mathrm A}\):
\begin{equation}
    (\mathbf V_c,\mathbf V^\star_{f_{1:H}})
    = \mathcal E_{\mathrm V}(O_c,\mathbf O^\star_{f_{1:H}}),
    \mathbf A^\star_{f_{1:H}}
    = \mathcal E_{\mathrm A}(\mathbf a^\star_{f_{1:H}}).
    \label{eq:training_target_encoding}
\end{equation}
Here, \(H\) denotes the prediction horizon and let \(\mathbf X^\star=(\mathbf V^\star_{f_{1:H}},\mathbf A^\star_{f_{1:H}})\). We sample random Gaussian noise \(\boldsymbol\epsilon=(\boldsymbol\epsilon_{\mathrm V},\boldsymbol\epsilon_{\mathrm A})\sim\mathcal N(\mathbf 0,\mathbf I)\) and flow time \(s\sim\mathcal U(0,1)\) to construct the interpolated target \(\mathbf X_s=(1-s)\boldsymbol\epsilon+s\mathbf X^\star\).

The MoT-style~\cite{liang2025mixtureoftransformers} dual-expert model $\mathcal F_\theta$ is trained through flow matching~\cite{lipman2023flow} with target velocity $\mathbf X^\star-\boldsymbol\epsilon$:
\begin{equation}
    \mathcal L_{\mathrm{FM}}
    =
    \mathbb E\!\left[
        \left\|
        \mathbf v_\theta(\mathbf X_s,s,\mathbf V_c,\mathbf Q_c,\mathbf T_l)
        -
        (\mathbf X^\star-\boldsymbol\epsilon)
        \right\|_2^2
    \right],
    \label{eq:flow_matching_loss}
\end{equation}
where $\mathbf v_\theta$ is the joint velocity predictor of $\mathcal F_\theta$, and the expectation is over training samples, flow time, and noise.

During inference, the model integrates the learned velocity field from Gaussian noise to generate future latents:
\begin{equation}
    (\mathbf V_{f_{1:H}},\mathbf A_{f_{1:H}})
    =
    \mathcal F_\theta\!\left(
        \mathbf V_c,\mathbf Q_c,\mathbf T_l,
        \boldsymbol\epsilon_{\mathrm V},
        \boldsymbol\epsilon_{\mathrm A}
    \right).
    \label{eq:joint_future_prediction}
\end{equation}
The VAE decoder $\mathcal D_{\mathrm V}$ and action decoder $\mathcal D_{\mathrm A}$ recover future observations and numerical robot actions:
\begin{equation}
    \mathbf O_{f_{1:H}}
    = \mathcal D_{\mathrm V}(\mathbf V_{f_{1:H}}),
    \qquad
    \hat{\mathbf a}_{f_{1:H}}
    = \mathcal D_{\mathrm A}(\mathbf A_{f_{1:H}}).
    \label{eq:output_decoding}
\end{equation}

\noindent\textbf{Network architecture.}
As shown in \hyperref[fig:method]{Fig.~\ref*{fig:method}(a)}, compared with existing WAMs, our SVA-WAM introduces long-horizon subgoal learning (LSL) and visual-action bridging (VAB). SVA-WAM adopts a MoT-style~\cite{liang2025mixtureoftransformers} three-expert architecture with a visual-action expert between the video and action experts. In LSL, both the video and visual-action experts predict long-horizon subgoals to guide their short-horizon predictions of visual dynamics and visual-action representations, respectively. In VAB, we render visual-action representations from the current robot state, the URDF file, and the link meshes. Similar to visual dynamics prediction, the visual-action expert then predicts future visual-actions conditioned on long-horizon visual-action subgoals. These future visual-action sequence guides the prediction of the final robot actions. In addition, we construct a structured attention mask to control how long-horizon prediction guides short-horizon prediction and how visual-action bridge visual and action representations.

\subsection{Long-Horizon Subgoal Learning}
WAMs jointly predict visual dynamics and robot actions and show superior generalization to visual perturbations. However, most existing WAMs predict only short-horizon visual dynamics without long-horizon subgoal guidance. Long-horizon subgoals can guide action prediction toward task completion, especially in long-horizon tasks. Moreover, these subgoals depend only on the task and are unaffected by task-irrelevant visual changes. Thereby, they can further improve generalization to visual perturbations. Based on this, we introduce long-horizon subgoal learning to guide short-horizon prediction.

To provide task-relevant long-horizon subgoal guidance, we first generate visual subgoal sequences from expert demonstrations. As illustrated in \hyperref[fig:method]{Fig.~\ref*{fig:method}(c)}, we divide each trajectory into segments based on task stages and annotate the final frame of each segment as a subgoal. This yields $K$ visual subgoals $\{G_i\}_{i=1}^{K}$ with corresponding frame indices $\mathcal{T}=\{t_1,t_2,\ldots,t_K\}$. We then introduce long-horizon subgoal learning. Given the current observation $O_c$, we use the first $N$ annotated subgoals after $c+H$ as supervision: $\{\mathcal{G}_n\}_{n=1}^{N}=\{O_{\tau_n}\mid \tau_n\in\mathcal{T},\ c+H\leq\tau_1<\cdots<\tau_N\leq T,\ n=1,\ldots,N\}$, where $T$ denotes the final frame index of the video. As in short-horizon prediction, long-horizon subgoal prediction is conditioned on the current observation latent $\mathbf V_c$ and instruction representation $\mathbf T_l$. We encode the annotated subgoal sequence using the same VAE encoder as for short-horizon visual dynamics: 
\begin{equation}
    (*,\mathbf V_{g_{n}})
    =
    \mathcal E_{\mathrm V}(O_c,\mathcal G_n),
    \label{eq:subgoal_encoding}
\end{equation}
where $\mathbf V_{g_n}$ denotes the latent representation of the $n$-th selected subgoal. We train the video expert to predict long-horizon subgoals using the flow-matching objective in Eq.~\eqref{eq:flow_matching_loss}, with $\mathbf V_{g_{1:n}}-\boldsymbol\epsilon_{\mathrm V}$ as the target velocity. As shown in \hyperref[fig:method]{Fig.~\ref*{fig:method}(b)}, we allow short-horizon visual dynamics features $\mathbf V_{f_{1:H}}$ to attend to the features of the first predicted subgoal $\mathbf V_{g_1}$, providing long-horizon guidance.

\subsection{Visual-Action Bridging}
Most existing WAMs rely only on predicted visual dynamics or other information about the visual scene to guide robot action generation. Visual dynamics contain task-relevant information and also include task-irrelevant information, such as backgrounds and distractors. Such irrelevant information may interfere with action prediction, thereby reducing the effectiveness of visual dynamics guidance. Although we introduce long-horizon subgoal learning in the previous section, these subgoals are still visual representations and may contain irrelevant information. To better transfer task-relevant information from visual dynamics to robot actions, we introduce visual-action representations.

As shown in \hyperref[fig:method]{Fig.~\ref*{fig:method}(d)}, visual-action are constructed from robot joint states $q_c$, link meshes $M$, and the URDF-defined kinematic structure $U$. Forward kinematics determines the pose of each link for rendering:
\begin{equation}
    \mathcal{VA}_c
    =\left\{\mathcal R\!\left(U,M,C_j,q_c\right)\right\}_{j=1}^{J},
    \label{eq:visual_action_rendering}
\end{equation}
where $C_j$ is the camera pose for viewpoint $j$. We use $J$ fixed orthogonal viewpoints that capture the entire robot and provide complementary spatial information, without explicitly aligning them with real camera viewpoints. These complementary views reduce single-view occlusion and projection ambiguities, preserving spatial pose information independently of scene appearance and camera setup.

As shown in \hyperref[fig:method]{Fig.~\ref*{fig:method}(a)}, we use the visual observation VAE encoder to encode visual-action for the current state, long-horizon subgoals, and short-horizon future dynamics, as in Eqs.~\eqref{eq:condition_encoding},~\eqref{eq:training_target_encoding}, and~\eqref{eq:subgoal_encoding}, yielding \(\mathbf{VA}_{c}\), \(\mathbf{VA}_{f_{1:H}}\), and \(\mathbf{VA}_{g_{1:n}}\). To bridge visual dynamics and robot actions, we introduce a visual-action expert between the video expert and action expert. This expert is trained to predict long-horizon visual-action subgoals and short-horizon visual-action sequences using the same flow-matching objective as in Eq.~\eqref{eq:flow_matching_loss}. As with visual dynamics, the long-horizon visual-action subgoals guide short-horizon visual-action prediction. As shown in \hyperref[fig:method]{Fig.~\ref*{fig:method}(b)}, the visual and visual-action modalities interact over during long-horizon prediction, while the visual-action and action modalities interact over the short horizon, forming a visual-action bridge.

\subsection{Layer-Grouped Cross-Modal Attention}
As discussed above, visual observations contain both task-relevant information and task-irrelevant distractors. To address this, we build an intermediate visual-action representation that provides task-relevant information to the action expert. To support this bridging, we design a structured, layer-grouped attention mask.

As illustrated in \hyperref[fig:method]{Fig.~\ref*{fig:method}(b)}, we split the three experts into shallow and deep groups by block depth. In the shallow group, only visual-action to visual attention is enabled (yellow blocks); in the deep group, this attention is disabled and action to visual-action attention is enabled (dark green blocks). This design reduces interference between modalities and provides the action expert with clear visual-action information for predicting robot actions.

\subsection{Training and Inference}
\noindent\textbf{Training.}
We train three experts in SVA-WAM through flow matching. The visual and visual-action losses each sum short-horizon and long-horizon subgoal prediction terms, while the action loss supervises action generation:
\begin{equation}
    \mathcal L
    =
    \lambda_{\mathrm V}\mathcal L_{\mathrm{FM}}^{\mathrm V}
    +
    \lambda_{\mathrm{VA}}\mathcal L_{\mathrm{FM}}^{\mathrm{VA}}
    +
    \lambda_{\mathrm A}\mathcal L_{\mathrm{FM}}^{\mathrm A},
    \label{eq:joint_training_objective}
\end{equation}
where the $\lambda$ coefficients balance the three losses, and both $\mathcal L_{\mathrm{FM}}^{\mathrm V}$ and $\mathcal L_{\mathrm{FM}}^{\mathrm{VA}}$ consist of short-horizon prediction and long-horizon subgoal prediction losses.

\noindent\textbf{Inference.}
At inference time, SVA-WAM skips short-horizon visual dynamics prediction and generates only the first visual subgoal, the first visual-action subgoal, future visual-action and the robot action. We apply partial denoising to the subgoal tokens and the future visual–action tokens, reducing inference cost while retaining the conditioning information needed for action generation.
\begin{table}[t]
\centering
\footnotesize
\caption{\textbf{Performance comparison on LIBERO.}
Success rates (\%) are reported for the Spatial, Object, Goal, and Long suites. Average denotes the mean across the four suites. P.T. indicates pretraining on embodied datasets. The best and second-best results in each column are shown in bold and underlined, respectively.}
\label{tab:libero}
\setlength{\tabcolsep}{4.5pt}
\begin{tabularx}{\linewidth}{l|c|cccc|c}
\toprule
\textbf{Method}
& \textbf{P.T.}
& \textbf{Spatial}
& \textbf{Object}
& \textbf{Goal}
& \textbf{Long}
& \textbf{Avg.} \\
\midrule

Fast-WAM~\cite{yuan2026fastwam}
& \xmark
& 98.2
& \textbf{100.0}
& 97.0
& 95.2
& 97.6 \\

Motus~\cite{bi2025motus}
& \cmark
& 96.8
& \underline{99.8}
& 96.6
& 97.6
& 97.7 \\

MaskWAM~\cite{yu2026maskwam}
& \xmark
& 98.8
& \textbf{100.0}
& 98.2
& 96.4
& 98.4 \\

LingBot-VA~\cite{li2026lingbotva}
& \cmark
& 98.5
& 99.6
& 97.2
& \textbf{98.5}
& 98.5 \\

SG-WAM~\cite{zhao2026sgwam}
& \xmark
& \underline{99.4}
& \underline{99.8}
& 98.6
& 96.2
& 98.5 \\

4D-WAM~\cite{yang2026fourdwam}
& \xmark
& \textbf{99.5}
& 98.6
& 98.8
& 97.4
& 98.6 \\

GeoSem-WAM~\cite{ma2026geosemwam}
& \xmark
& 99.0
& \textbf{100.0}
& 98.2
& 97.0
& 98.6 \\

ST-WAM~\cite{wang2026stwam}
& \xmark
& 99.0
& \textbf{100.0}
& \underline{99.0}
& 96.8
& \underline{98.7} \\

\textbf{SVA-WAM (Ours)}
& \xmark
& 98.0
& 99.4
& \textbf{99.6}
& \underline{98.2}
& \textbf{98.8} \\

\bottomrule
\end{tabularx}
\end{table}
\begin{table*}[t]
\centering
\footnotesize
\caption{\textbf{Zero-shot performance comparison on LIBERO-Plus.}
Success rates (\%) are reported across seven evaluation conditions, with their weighted mean shown in the last column. P.T. indicates pretraining on embodied datasets. The best and second-best results within each model group are shown in bold and underlined, respectively.}
\label{tab:libero_plus}

\setlength{\tabcolsep}{7.4pt}
\begin{tabularx}{\linewidth}{l|c|c|ccccccc|c}
\toprule
\textbf{Method} & \textbf{Publication}
& \textbf{P.T.} & \textbf{Camera} & \textbf{Robot} & \textbf{Language} & \textbf{Light}
& \textbf{Background} & \textbf{Noise} & \textbf{Layout} & \textbf{Avg.} \\
\midrule

\multicolumn{11}{l}{\textit{\textbf{Vision-Language-Action models (VLAs)}:}} \\
\midrule

OpenVLA~\cite{kim2024openvla} & CoRL 2024
& \cmark & 0.8 & 3.5 & 23.0 & 8.1
& 34.8 & 15.2 & 28.5 & 15.6 \\

WorldVLA~\cite{cen2025worldvla} & arXiv 2025
& \cmark & 0.1 & 27.9 & 41.6 & 43.7
& 17.1 & 10.9 & 38.0 & 25.0 \\

UniVLA~\cite{bu2025univla} & RSS 2025
& \cmark & 1.8 & \underline{46.2} & 69.6 & 69.0
& 81.0 & 21.2 & 31.9 & 42.9 \\

$\pi_0$~\cite{black2024pi0} & RSS 2025
& \cmark & 13.8 & 6.0 & 58.8 & 85.0
& 81.4 & \textbf{79.0} & 68.9 & 53.6 \\

$\pi_0$-Fast~\cite{pertsch2025fast} & RSS 2025
& \cmark & \textbf{65.1} & 21.6 & 61.0 & 73.2
& 73.2 & 74.4 & 68.8 & 61.6 \\

OpenVLA-OFT~\cite{kim2025openvlaoft} & RSS 2025
& \cmark & \underline{56.4} & 31.9
& \textbf{79.5} & \textbf{88.7}
& \underline{93.3} & \underline{75.8}
& \textbf{74.2} & \underline{69.6} \\

X-VLA~\cite{zheng2025xvla} & ICLR 2026
& \cmark & 23.4 & \textbf{89.7}
& \underline{75.7} & \underline{88.2}
& \textbf{96.0} & 62.7
& \underline{71.8} & \textbf{70.5} \\

\midrule
\multicolumn{11}{l}{\textit{\textbf{World Action Models (WAMs):}}} \\
\midrule

Fast-WAM~\cite{yuan2026fastwam} & NeurIPS 2026
& \xmark & 21.0 & 49.5 & 73.3 & 77.5
& 52.0 & 46.7 & 63.7 & 53.8 \\

4D-WAM~\cite{yang2026fourdwam} & arXiv 2026
& \xmark & 45.2 & \underline{64.3}
& \textbf{90.6} & \underline{94.3}
& 57.7 & 69.1 & \textbf{79.2} & 71.0 \\

GaussianWAM~\cite{zhang2026gaussianwam} & arXiv 2026
& \xmark & 54.7 & 61.6 & 75.4 & 90.8
& 58.2 & \textbf{83.0} & \underline{76.7} & 71.3 \\

ST-WAM~\cite{wang2026stwam} & arXiv 2026
& \xmark & \underline{55.4} & 60.1 & 79.3 & 93.0
& \underline{74.2} & 79.5
& 74.3 & 72.8 \\

SG-WAM~\cite{zhao2026sgwam} & arXiv 2026
& \xmark & \textbf{58.6} & 48.9 & 81.4 & 89.8
& \textbf{86.1} & \underline{80.7} & 74.2 & \underline{73.0} \\

\textbf{SVA-WAM (Ours)} & ---
& \xmark & 52.6 & \textbf{73.9}
& \underline{88.0} & \textbf{94.4}
& 69.2 & 75.0 & 74.6 & \textbf{74.8} \\

\rowcolor{black!6}
\textit{$\Delta$ vs.\ baseline} & &
& \textcolor{red}{\textit{+31.6}}
& \textcolor{red}{\textit{+24.4}}
& \textcolor{red}{\textit{+14.7}}
& \textcolor{red}{\textit{+16.9}}
& \textcolor{red}{\textit{+17.2}}
& \textcolor{red}{\textit{+28.3}}
& \textcolor{red}{\textit{+10.9}}
& \textcolor{red}{\textit{+21.0}} \\

\bottomrule
\end{tabularx}
\end{table*}

\section{Experiments}

\subsection{Experimental Setup}

\noindent\textbf{Datasets.}
We evaluate SVA-WAM on three simulation benchmarks: LIBERO~\cite{liu2023libero}, LIBERO-Plus~\cite{fei2026liberoplus}, and RMBench~\cite{chen2026rmbench}. LIBERO comprises four task suites with 40 tasks in total. LIBERO-Plus extends the original LIBERO tasks with seven types of perturbations. We evaluate the model trained on LIBERO directly on LIBERO-Plus in a zero-shot setting, without additional fine-tuning. RMBench is a simulation benchmark for long-horizon tasks, comprising nine tasks with 50 demonstration trajectories each. 
For real-world evaluation, we use the Mirror robot platform and design three tasks: (1) \textbf{Place Cup:} places cups in designated rack slots in a specified color order. (2) \textbf{Build Block:} stacks blocks into a tower in a specified color order. (3) \textbf{Tidy Tableware:} puts plates and cups in their racks and adds bowls to an existing stack. We collect approximately 200 demonstrations per task. 

\noindent\textbf{Implementation details.}
We initialize the video expert with pretrained Wan2.2-5B~\cite{wan2025} weights. 
The visual-action and action experts adopt the same architecture as the video expert, except for a hidden-state dimension of 1024, and are initialized through parameter interpolation from its pretrained weights. We freeze the VAE encoder and text encoder, training only the three experts and the action encoder and decoder. To learn high-quality subgoal and visual-action predictions that guide action generation, we adopt a two-stage training strategy. In the first stage, we train only the video and visual-action experts to predict both dense short-horizon sequences and sparse long-horizon subgoals in their respective modalities. In the second stage, we jointly fine-tune all trainable components. More experimental details are presented in ‌Supplementary Material.

\subsection{Simulation Experiments}

\noindent\textbf{On LIBERO.}
In Tab.~\ref{tab:libero}, SVA-WAM achieves an average success rate of 98.8\% on LIBERO~\cite{liu2023libero}, surpassing baseline Fast-WAM~\cite{yuan2026fastwam} by 1.2\%. The improvement is particularly notable on the Long suite, where SVA-WAM achieves a success rate of 98.2\%, outperforming the baseline by 3.0\%. This result highlights the effectiveness of our framework in long-horizon task execution.

\noindent\textbf{On LIBERO-Plus.}
Tab.~\ref{tab:libero_plus} compares our SVA-WAM with VLA and WAM methods on LIBERO-Plus~\cite{fei2026liberoplus}. SVA-WAM achieves an average success rate of 74.8\% under zero-shot evaluation, exceeding baseline Fast-WAM~\cite{yuan2026fastwam} by 21.0\%, with improvements across all seven perturbation categories. It also compares favorably with other VLA and WAM methods, outperforming X-VLA~\cite{zheng2025xvla} and SG-WAM~\cite{zhao2026sgwam} by 4.3\% and 1.8\%, respectively. The gains across perturbation categories may stem from long-horizon subgoals that keep short-horizon predictions aligned with task milestones, and from visual-action representations that encode robot motion independently of scene appearance.

\begin{figure}[t]
\centering
\includegraphics[width=1.0\linewidth]{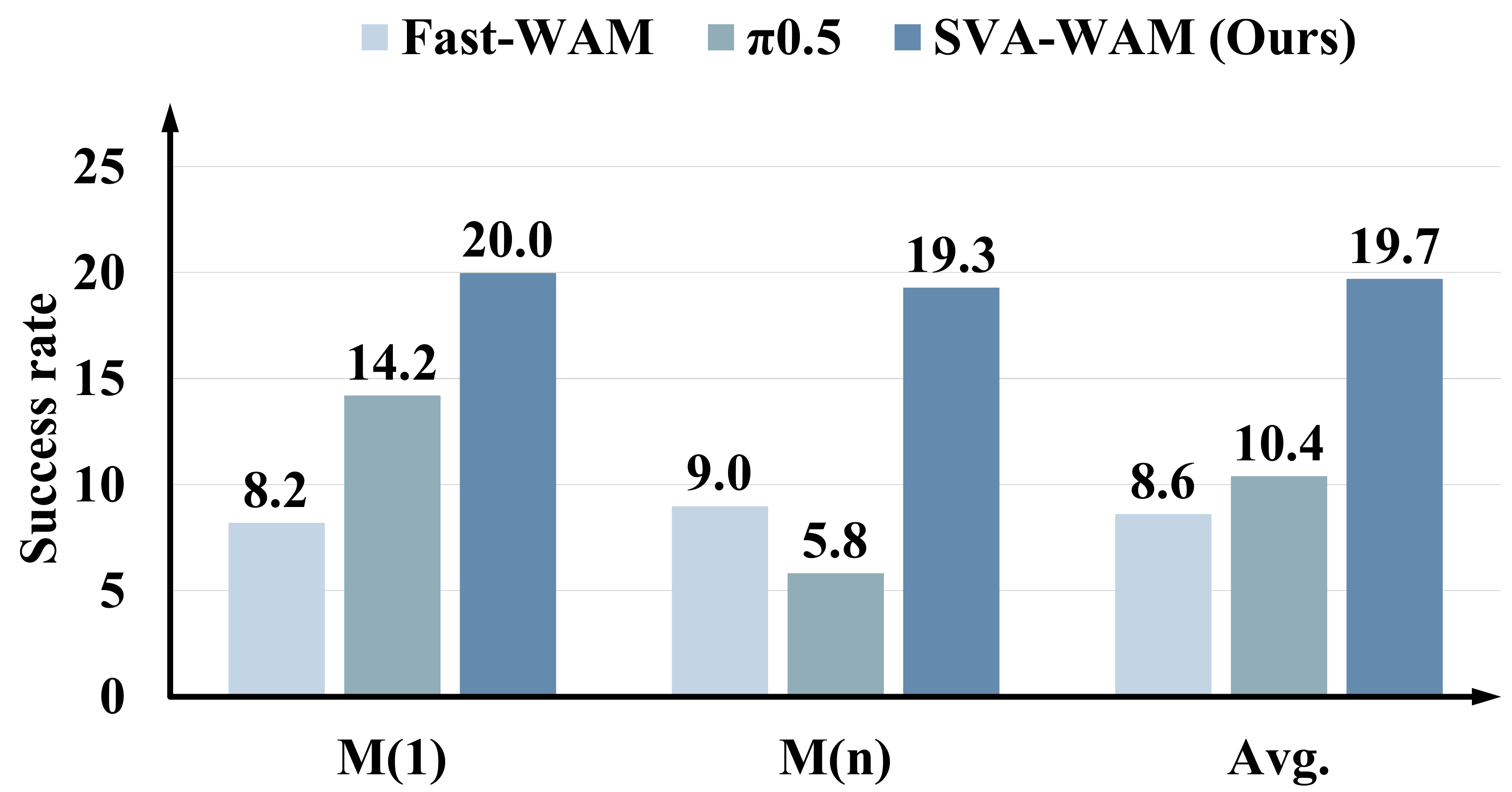}
\caption{\textbf{Performance comparison on RMBench.}
SVA-WAM achieves higher mean success rates in both $M(1)$ and $M(n)$. Avg.\ denotes the mean success rate across all RMBench tasks.}

\label{fig:rmbench}
\end{figure}

\noindent\textbf{On RMBench.}
As shown in Fig.~\ref{fig:rmbench}, SVA-WAM achieves an overall success rate of 19.7\% on RMBench~\cite{chen2026rmbench} benchmark, outperforming Fast-WAM~\cite{yuan2026fastwam} and $\pi_{0.5}$~\cite{black2025pi05} by 11.1\% and 9.3 \%, respectively. Under the $M(1)$ and $M(n)$ settings, SVA-WAM reaches 20.0\% and 19.3\%, exceeding Fast-WAM by 11.8\% and 10.3\% and $\pi_{0.5}$ by 5.8\% and 13.5\%, respectively. These results suggest that SVA-WAM benefits from long-horizon subgoal guidance, enabling stronger performance on long-horizon manipulation tasks.

\subsection{Real-World Experiments}

\noindent\textbf{Overall performance.}
In Tab.~\ref{tab:real}, SVA-WAM achieves success rates of 83.3\%, 66.7\%, and 58.3\% on the Place Cup, Build Block and Tidy Tableware tasks, respectively, outperforming baseline Fast-WAM~\cite{yuan2026fastwam} by 25.0\%, 41.7\%, and 41.6\%. It also surpasses $\pi_{0.5}$~\cite{black2025pi05} by 8.3\%, 25.0\% and 16.6\% on three tasks. Furthermore, SVA-WAM attains the highest progress scores across all three tasks, reaching  28.3, 33.3 and 60.1 out of maximum scores of 30, 40, and 70, respectively. These results demonstrate the effectiveness of our method on both the simple Place Cup task and the long-horizon complex  Build Block and Tidy Tableware tasks.

\begin{table}[t]
\centering
\footnotesize
\caption{\textbf{Real-world task performance.}
We compare SVA-WAM with Fast-WAM and $\pi_{0.5}$ using success rates (\%) and progress scores. The maximum progress scores for Place Cup, Build Block, and Tidy Tableware are 30, 40, and 70, respectively.}
\label{tab:real}

\setlength{\tabcolsep}{3.0pt}

\begin{tabularx}{\linewidth}{l|ccc|c}
\toprule
\textbf{Method}
& \textbf{Place Cup} & \textbf{Build Block} & \textbf{Tidy Tableware} & \textbf{Avg.} \\
\midrule
\multicolumn{5}{l}{\textit{\textbf{Success Rate:}}} \\
\midrule
Fast-WAM
& 58.3 & 25.0 & 16.7 & 33.3 \\
$\pi_{0.5}$
& 75.0 & 41.7 & 41.7 & 52.8 \\
\textbf{SVA-WAM (Ours)}
& \textbf{83.3} & \textbf{66.7} & \textbf{58.3} & \textbf{69.4} \\
\midrule
\multicolumn{5}{l}{\textit{\textbf{Progress Score:}}} \\
\midrule
Fast-WAM
& 22.5 & 19.2 & 28.3 & 23.3 \\
$\pi_{0.5}$
& 27.5 & 28.3 & 57.5 & 37.8 \\
\textbf{SVA-WAM (Ours)}
& \textbf{28.3} & \textbf{33.3} & \textbf{60.1} & \textbf{40.6} \\
\bottomrule
\end{tabularx}
\end{table}

\begin{figure}[t]
\centering
\includegraphics[width=0.95\linewidth]{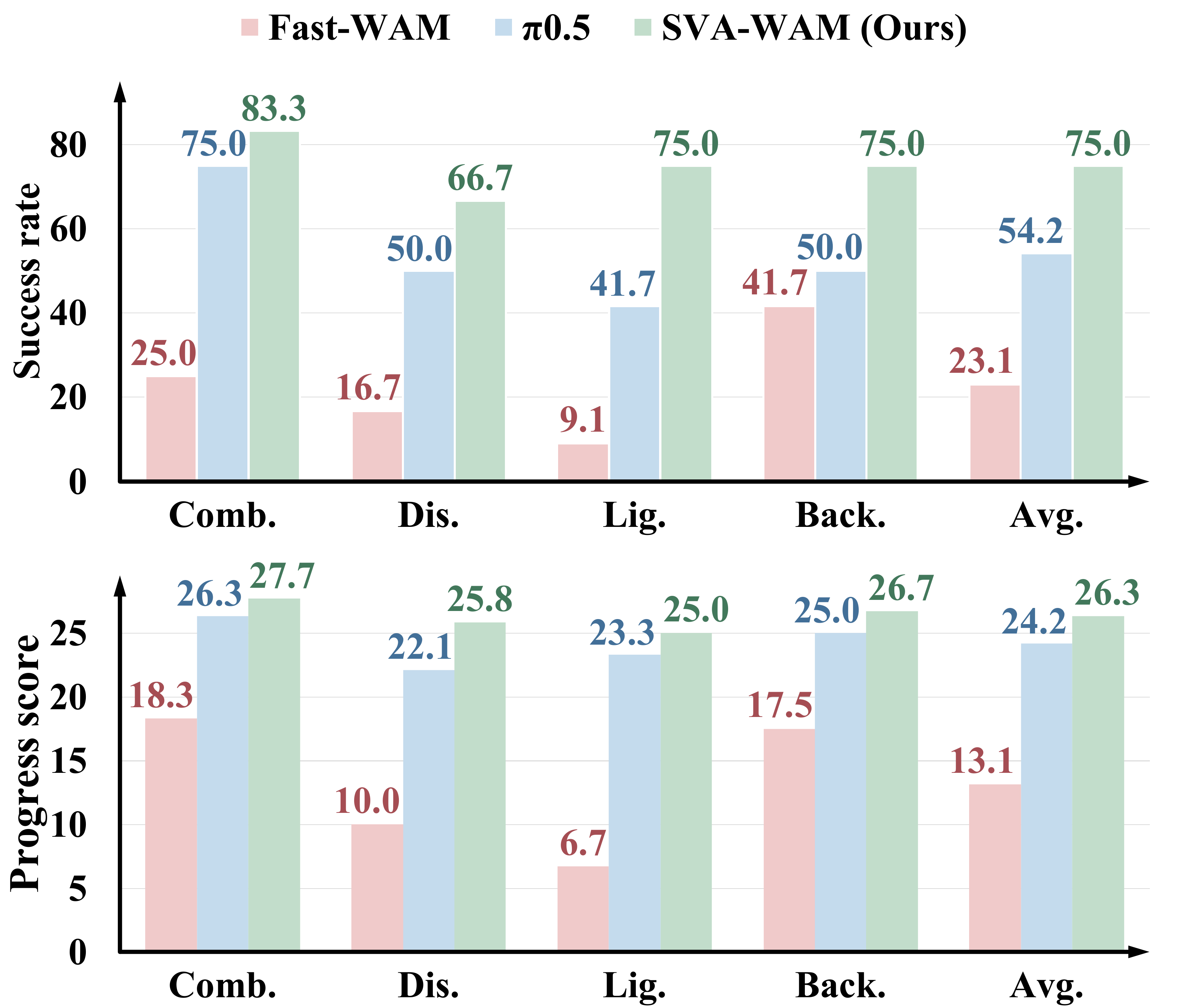}
\caption{\textbf{Real-world generalization results.}
We compare performance on the Place Cup task under four out-of-distribution perturbations: \textbf{Comb.}, unseen combinations of placement orders and positions; \textbf{Dis.}, distractors absent from the demonstrations; \textbf{Lig.}, altered lighting; and \textbf{Back.}, a different tabletop background. We report success rates and progress scores respectively. }

\label{fig:gen}
\end{figure}

\noindent\textbf{Generalization tests.}
We evaluate generalization of our SVA-WAM on the Place Cup task under four out-of-distribution perturbations: unseen combinations of placement orders and positions, distractors absent from the demonstrations, altered lighting, and a different tabletop background. In Fig.~\ref{fig:gen}, SVA-WAM achieves an average success rate of 75.0\% across the four test conditions, surpassing Fast-WAM~\cite{yuan2026fastwam} and $\pi_{0.5}$~\cite{black2025pi05} by 51.9\% and 20.8\%, respectively. It also achieves the highest average progress score of 26.3, compared with 13.1 for Fast-WAM and 24.2 for $\pi_{0.5}$. These results show that our SVA-WAM is robust to visual variations in real-world tasks.

\begin{figure}[t]
\centering
\includegraphics[width=1.0\linewidth]{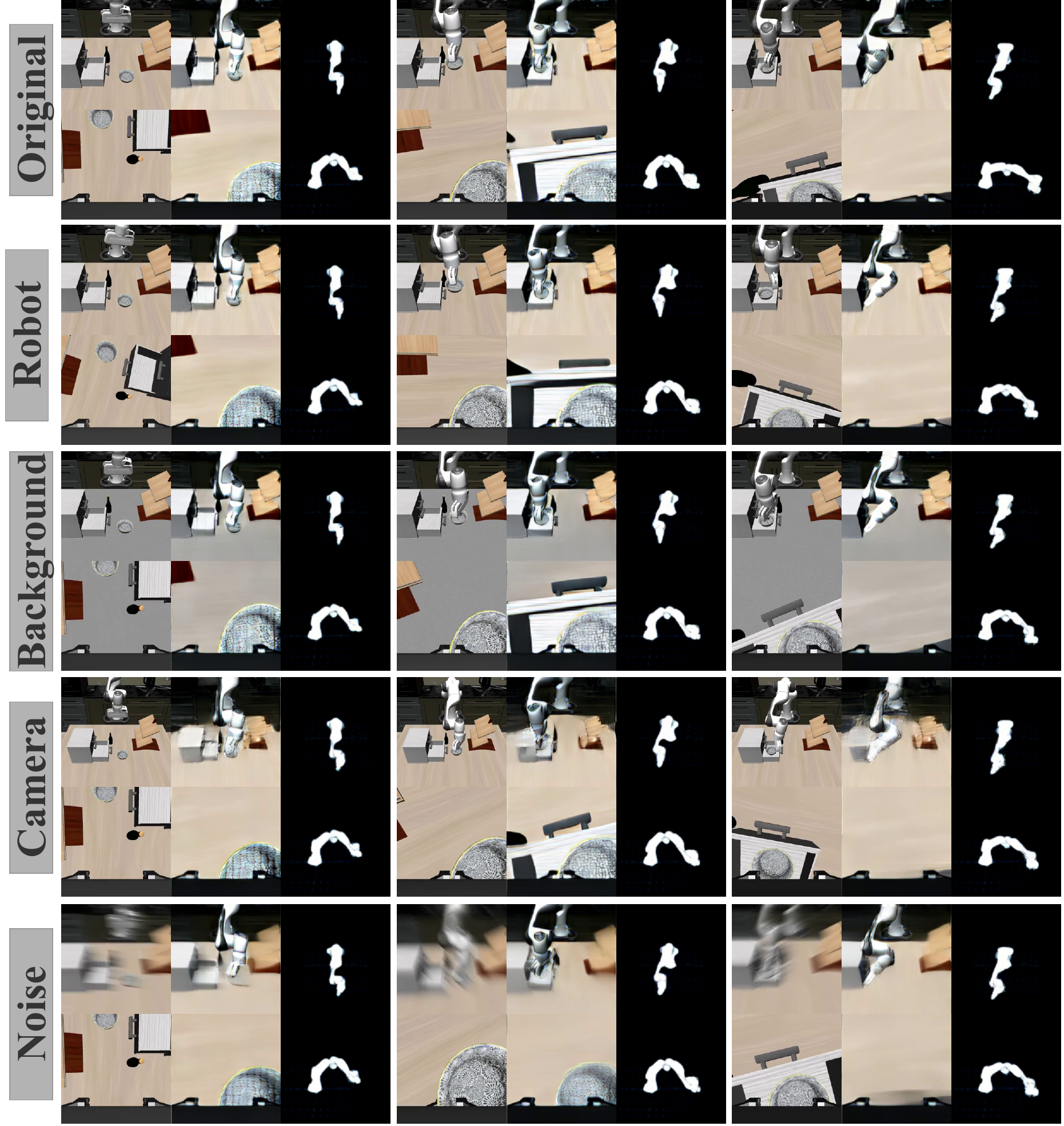}
\caption{
\textbf{Visualization of model predictions.}
Rows show the original setting and robot, background, camera, and noise perturbations; columns show three examples.
Each panel presents the input images, predicted visual subgoals, and predicted visual-action subgoals from left to right.
The first two use main and wrist views, while visual-action use two orthogonal views.
}
\label{fig:result_vis}
\end{figure}

\subsection{Qualitative Results}
Fig.~\ref{fig:result_vis} shows the first visual and visual-action subgoals predicted by our SVA-WAM to guide action generation during LIBERO~\cite{liu2023libero} and LIBERO-Plus~\cite{fei2026liberoplus} evaluation. SVA-WAM predicts both subgoals accurately in in-domain tests and under several scene perturbations. For instance, when the robot starts from different poses, it predicts similar subgoals because the overall task goal remains the same. Camera viewpoint changes and image noise reduce the quality of visual subgoal predictions. However, since the visual-action representation focuses on robot motion, the predicted visual-action subgoals remain relatively accurate and help guide correct action generation.

\subsection{Ablation Studies}

\noindent\textbf{Component ablations.}
Tab.~\ref{tab:ab_module} shows the contributions of the main components of our SVA-WAM. Removing subgoal guidance reduces success rates by 2.4\% and 9.7\% on LIBERO-Long~\cite{liu2023libero} and LIBERO-Plus~\cite{fei2026liberoplus}, respectively. This suggests that subgoal guidance supports both long-horizon task execution and generalization under environmental perturbations. Removing visual-actions leads to a modest drop of 0.5\% on LIBERO-Avg but a larger decline of 4.1\% on LIBERO-Plus, highlighting the contribution of robot-centric visual-action representations to generalization. Removing layer-grouped attention mask decreases LIBERO-Plus performance by 2.9\%, suggesting that this attention structure also improves robustness to environmental perturbations. Finally, removing the first stage of visual training reduces LIBERO-Plus performance by 8.3\%. This suggests that learning to predict subgoals and visual-action during the first training stage helps the model generalize by providing better guidance for action generation.

\begin{table}[t]
\centering
\footnotesize
\caption{\textbf{Component ablations on LIBERO and LIBERO-Plus.}
We evaluate the effects of removing long-horizon subgoal learning (LSL), visual-action bridging (VAB), layer-grouped attention mask (LM), and first-stage training (S1). Results are success rates (\%) on LIBERO-Long, averaged across the four LIBERO suites, and on LIBERO-Plus.}
\label{tab:ab_module}

\setlength{\tabcolsep}{6pt}

\begin{tabularx}{\linewidth}{l|ccc}
\toprule
\textbf{Method}
& \textbf{LIBERO-Long}
& \textbf{LIBERO-Avg}
& \textbf{LIBERO-Plus} \\
\midrule
\textbf{SVA-WAM}
& \textbf{98.2} & \textbf{98.8} & \textbf{74.8} \\
\quad \textit{w/o} LSL
& 95.8 & 98.2 & 65.1 \\
\quad \textit{w/o} VAB
& 96.6 & 98.3 & 70.7 \\
\quad \textit{w/o} LM
& 96.6 & 98.4 & 71.9 \\
\quad \textit{w/o} S1
& 97.0 & 98.1 & 66.5 \\
\bottomrule
\end{tabularx}
\end{table}

\begin{figure}[t]
\centering
\includegraphics[width=0.95\linewidth]{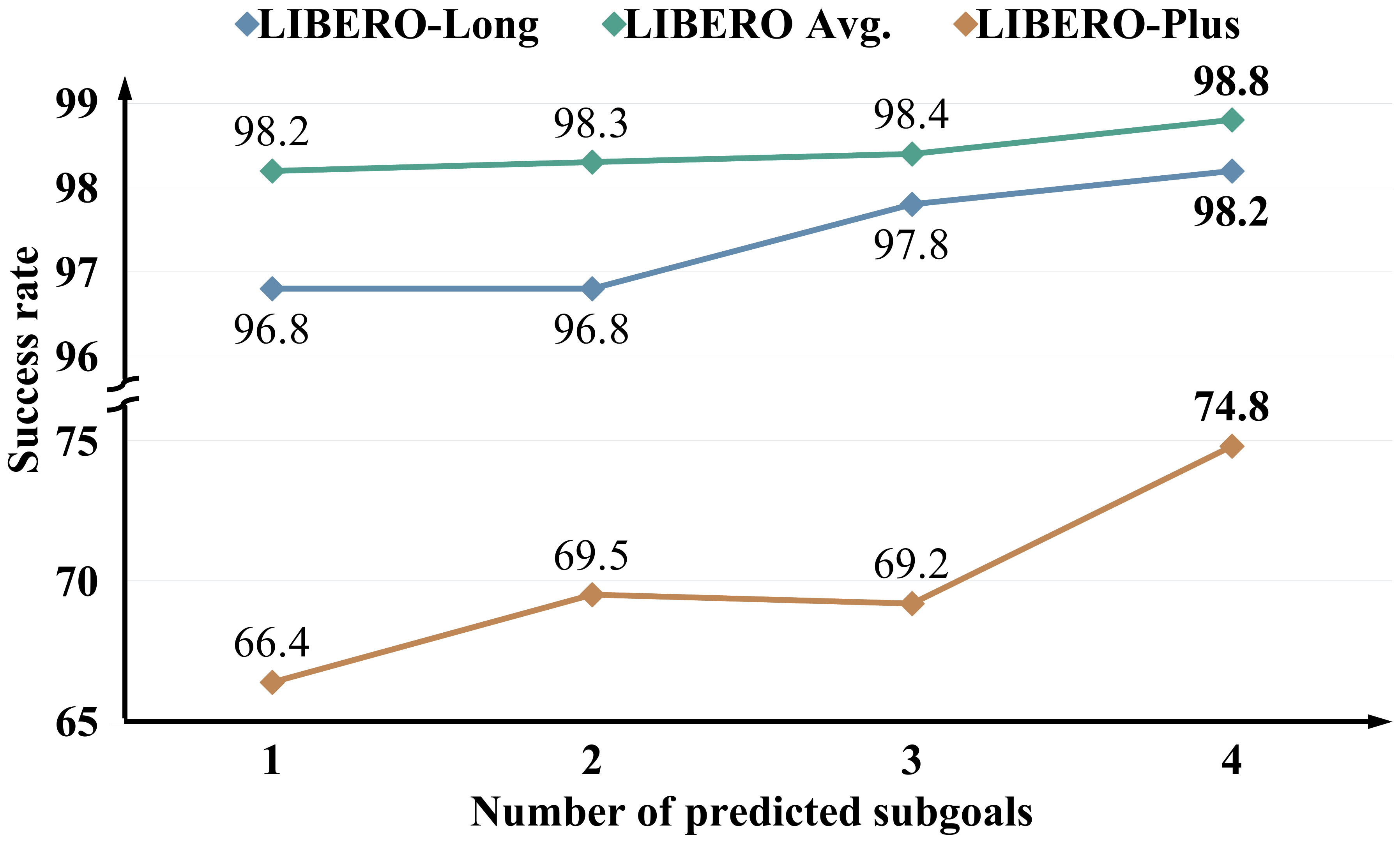}
\caption{\textbf{Effect of the number of subgoals predicted during training.} Success rates on LIBERO-Long, LIBERO-Avg., and LIBERO-Plus for models trained to predict one to four long-horizon visual subgoals. Predicting four subgoals during training yields the highest success rates across all three metrics.}
\label{fig:ab_num}
\end{figure}

\noindent\textbf{Number of subgoals predicted during training.}
Fig.~\ref{fig:ab_num}  shows how the number of subgoals predicted during training affects model performance. Increasing from one to four improves success rates from 96.8\% to 98.2\% on LIBERO-Long~\cite{liu2023libero}, from 98.2\% to 98.8\% on LIBERO-Avg, and from 66.4\% to 74.8\% on LIBERO-Plus~\cite{fei2026liberoplus}. Four subgoals yield the best results across all three metrics among the evaluated settings. Even though we use only the first predicted subgoal to guide action generation at inference, predicting multiple subgoals during training provides additional long-horizon supervision. This additional supervision may help the model predict a better first subgoal and thereby improve action generation. LIBERO-Plus performance remained nearly unchanged from two to three subgoals, possibly because most tasks have two or four subgoals, and three predictions still cannot fully cover 	tasks with four subgoals.

\begin{table}[t]
\centering
\footnotesize
\caption{\textbf{Design ablations on LIBERO and LIBERO-Plus.} We compare (a) two goal selection methods: subgoal and uniform sampling, (b) visual-action viewpoints at $0^\circ$ and $45^\circ$, and (c) two stage 1 training settings: visual (V) + visual-action (VA) and visual only (V). Success rates (\%) are reported for LIBERO-Long, the average across the four LIBERO suites, and LIBERO-Plus.}
\setlength{\tabcolsep}{3pt}

\resizebox{\linewidth}{!}{%
  \begin{tabular}{l|c|ccc}
    \toprule
    \multirow{3}{*}{(a)}
      & \textbf{Goal}
      & \textbf{LIBERO-Long}
      & \textbf{LIBERO-Avg.}
      & \textbf{LIBERO-Plus} \\
    \cmidrule{2-5}
      & Subgoal & \textbf{98.2} & \textbf{98.8} & \textbf{74.8} \\
      & Uniform & 97.8 & 98.6 & 68.3 \\

    \midrule
    \multirow{3}{*}{(b)}
      & \textbf{View angle}
      & \textbf{LIBERO-Long}
      & \textbf{LIBERO-Avg.}
      & \textbf{LIBERO-Plus} \\
    \cmidrule{2-5}
      & $0^\circ$  & \textbf{98.2} & \textbf{98.8} & \textbf{74.8} \\
      & $45^\circ$ & 97.8 & 98.5 & 73.1\\

    \midrule
    \multirow{3}{*}{(c)}
      & \textbf{Training S1}
      & \textbf{LIBERO-Long}
      & \textbf{LIBERO-Avg.}
      & \textbf{LIBERO-Plus} \\
    \cmidrule{2-5}
      & V + VA & \textbf{98.2} & \textbf{98.8} & \textbf{74.8} \\
      & V & 97.8 & \textbf{98.8} & 72.5 \\
    \bottomrule
  \end{tabular}
}

\label{tab:ablation_groups}
\end{table}
\noindent\textbf{Ablation study on different designs.}
Tab.~\ref{tab:ablation_groups} evaluates three design choices in SVA-WAM on LIBERO~\cite{liu2023libero} and LIBERO-Plus~\cite{fei2026liberoplus} benchmark. For goal selection, we replace subgoals with several uniformly sampled frames from a longer future horizon beyond the short-term segment. Uniform sampling achieves similar performance on LIBERO, but its success rate on LIBERO-Plus drops by 6.5\%. This gap may arise because subgoals at stage boundaries provide clearer targets for action prediction under visual changes. For viewpoint, the 0° viewpoint follows the real video’s viewing direction without exact alignment, while the 45° viewpoint is obtained by rotating the 0° viewpoint by 45°. This rotation slightly reduces performance on both LIBERO and LIBERO-Plus, suggesting that similar viewing directions are better. For the first training stage, we remove visual-action supervision and retain only visual supervision. This reduces the success rate on LIBERO-Plus by 2.3\%, suggesting that learning to predict visual-action in the first stage can better guide action generation.

\section{Conclusions}
We presented SVA-WAM, a framework for learning future representations for WAMs that combines LSL and VAB. The model bridges long-horizon prediction and short-horizon action generation while connecting visual dynamics representations with action representations. Experiments on simulation benchmarks and real-world robot tasks demonstrate our model's strong long-horizon task performance and generalization under visual variations.

\noindent\textbf{Future work.} Our framework has not yet been pretrained on embodied data, thereby its subgoal prediction and visual-action modeling capabilities have yet to be scaled up. Future work will explore how large-scale training with long-horizon future prediction and visual-action modeling enhances model capabilities.

\section*{LLM Usage Statement}
\label{sec:sup_llm}
We used Codex for language refinement, including grammar correction and wording improvements, as well as code assistance. The authors have carefully verified the correctness of the final content. 

{
    \small
    \bibliographystyle{main}
    \bibliography{main}
}
\clearpage
\setcounter{page}{1}
\appendix

\twocolumn[{%
  \centering
  {\Large\bfseries \thetitle\par}
  \vspace{0.5em}
  {\Large Supplementary Material\par}
  \vspace{1em}

\begin{minipage}{\textwidth}
  \centering
  \includegraphics[width=0.96\linewidth]{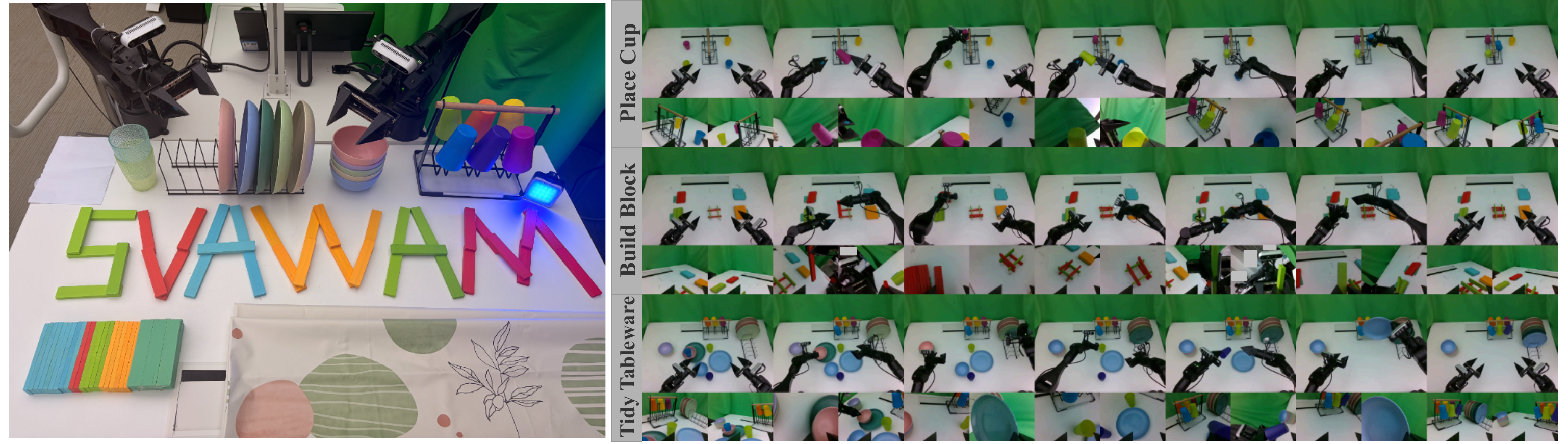}
  \captionof{figure}{\textbf{Real-world setup and task demonstrations.}
  The left panel shows the Mirror dual-arm tabletop robot and the objects
  used in the experiments. The right panel shows demonstration sequences
  for Place Cup, Build Block, and Tidy Tableware from overhead and
  wrist-mounted camera views.}
  \label{fig:sup_real_data}
\end{minipage}\par
  \vspace{1em}
}]

\section{Dataset Description}
\label{sec:sup_data}

\begin{figure}[t]
\centering
\includegraphics[width=0.95\linewidth]{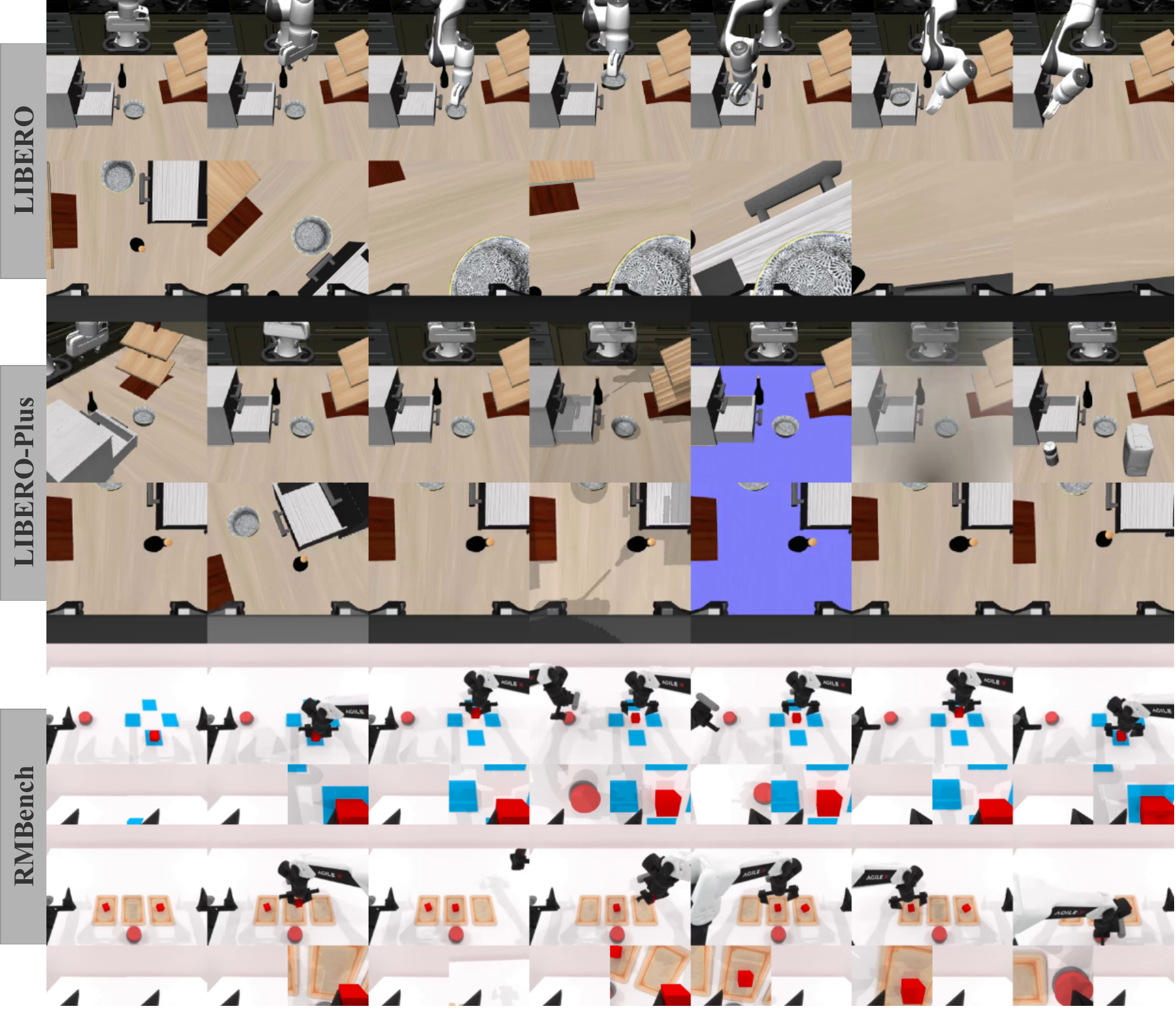}
\caption{\textbf{Example trajectories from simulation benchmarks.}
From top to bottom, the panels show manipulation sequences from LIBERO,
LIBERO-Plus, and RMBench. Frames within each sequence are ordered
chronologically from left to right.}

\label{fig:sup_data}
\end{figure}
\subsection{Real-World Datasets}
As shown in Fig.~\ref{fig:sup_real_data}, we conduct real-world experiments on the Mirror robot platform, a
dual-arm tabletop system equipped with three Intel RealSense D435
cameras. One overhead camera provides a global view of the workspace,
and two wrist-mounted cameras provide close-up views. We design three
real-world tasks:

\begin{itemize}
    \item \textbf{Place Cup.} The robot places the cups at designated
    locations in the color order specified by the instruction.
    For example, ``\textit{The positions on the cup rack are numbered one to six from left to right and top to bottom. Place the cups on the rack in the following order: first, place the red cup in position one; then, place the green cup in position three; finally, place the blue cup in position five.}''

    \item \textbf{Build Block.} The robot builds a four-layer block
    tower by stacking blocks in the color order specified by the
    instruction. For example, ``\textit{Build a four-layer square tower using blocks, following the color sequence: red, green, yellow, blue.}''

    \item \textbf{Tidy Tableware.} The robot moves each piece of
    tableware on the tabletop to its designated location. For example, ``\textit{Clear the tableware from the table, put the plates on the rack, place the cups upside down on the rack, and stack the bowls.}''
\end{itemize}
\subsection{Simulation Datasets}
\noindent\textbf{LIBERO.} The top row of Fig.~\ref{fig:sup_data} shows example trajectories from LIBERO. The benchmark uses a Franka robot arm and comprises four suites of 10 tasks each. LIBERO-Spatial varies object layouts, LIBERO-Object varies the objects to manipulate, and LIBERO-Goal varies task goals. LIBERO-Long focuses on multi-step manipulation.

\noindent\textbf{LIBERO-Plus.} LIBERO-Plus extends LIBERO to evaluate generalization under seven types of input and environmental perturbations. The second row of Fig.~\ref{fig:sup_data} shows examples from the benchmark. The perturbations modify object layouts by adding distractors or moving target objects, change camera position, orientation, or field of view, and alter the robot arm's initial configuration. They also rephrase task instructions, vary lighting intensity, direction, or color, replace scene and work-surface textures, and introduce sensor noise such as image blur.

\noindent\textbf{RMBench.} The third row of Fig.~\ref{fig:sup_data} shows example trajectories from RMBench, a simulation benchmark for long-horizon dual-arm manipulation. The five $M(1)$ tasks rely primarily on information from a single key observation and include Observe and Pick Up, Put Back Block, Rearrange Blocks, Swap Blocks, and Swap T. The four $M(n)$ tasks require integrating information across multiple steps through exploration, trial and error, or counting. These tasks include Battery Try, Blocks Ranking Try, Cover Blocks, and Press Button. The original study provides 50 expert demonstrations per task and evaluates each task over 100 episodes.

\begin{figure}[t]
\centering
\includegraphics[width=1.0\linewidth]{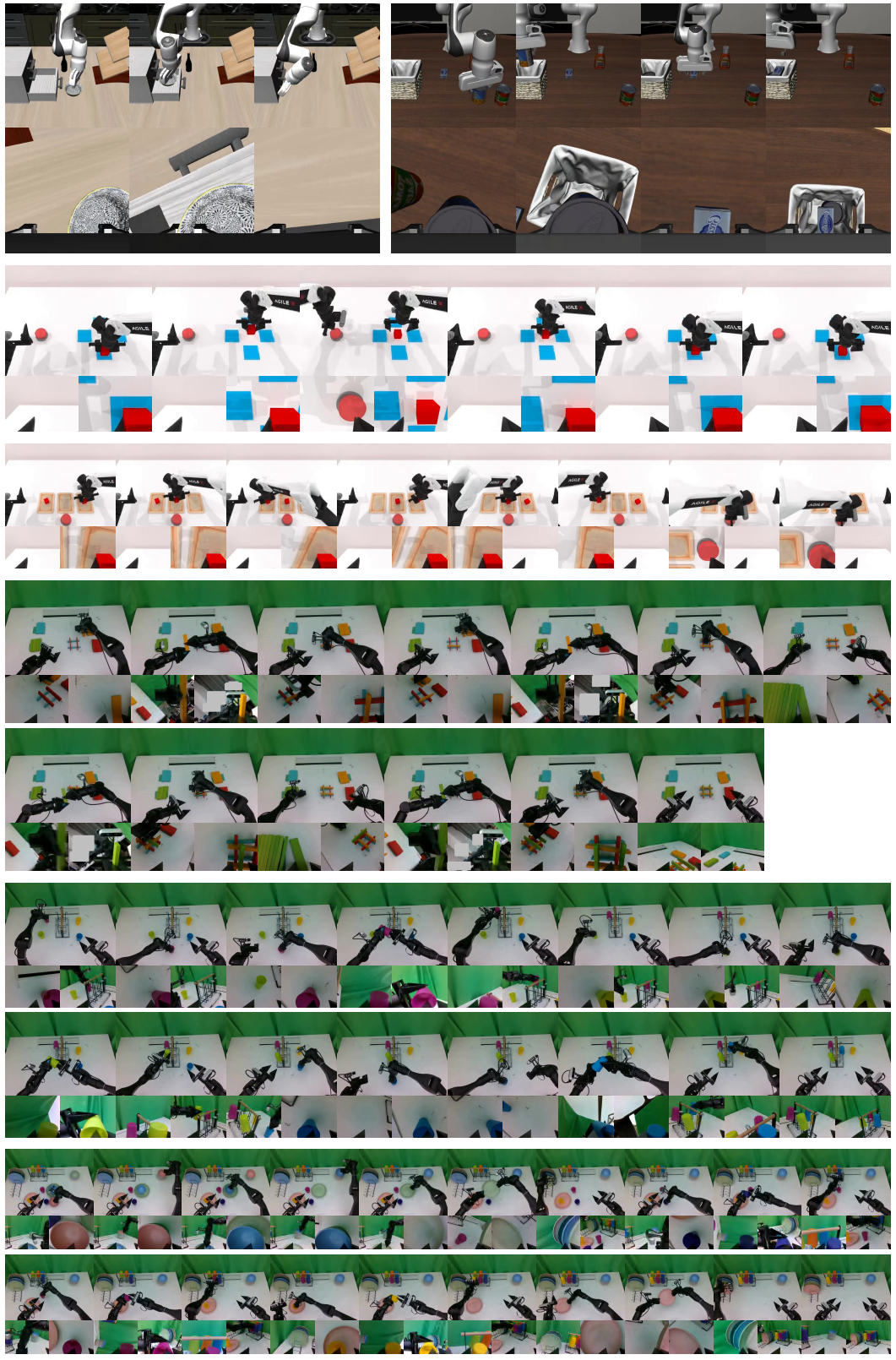}
\caption{\textbf{Example visual subgoal sequences.} From top to bottom, the figure shows examples from LIBERO, RMBench, and real-world tasks. Within each sequence, the selected subgoal frames appear in chronological order from left to right.}
\label{fig:sup_goal}
\end{figure}

\subsection{Subgoal Generation}
To construct visual subgoal sequences, we first identify frames in the expert demonstrations where the gripper opens or closes, as these events often coincide with robot--object interactions. We then manually review these candidate frames and retain those that mark the completion of a
subtask. We include the last frame of each demonstration video as the
final subgoal. For some shorter tasks, we use this frame as the sole
subgoal. Fig.~\ref{fig:sup_goal} shows example visual subgoals for tasks from LIBERO and RMBench, as well as for real-world tasks.

\begin{figure}[t]
\centering
\includegraphics[width=1.0\linewidth]{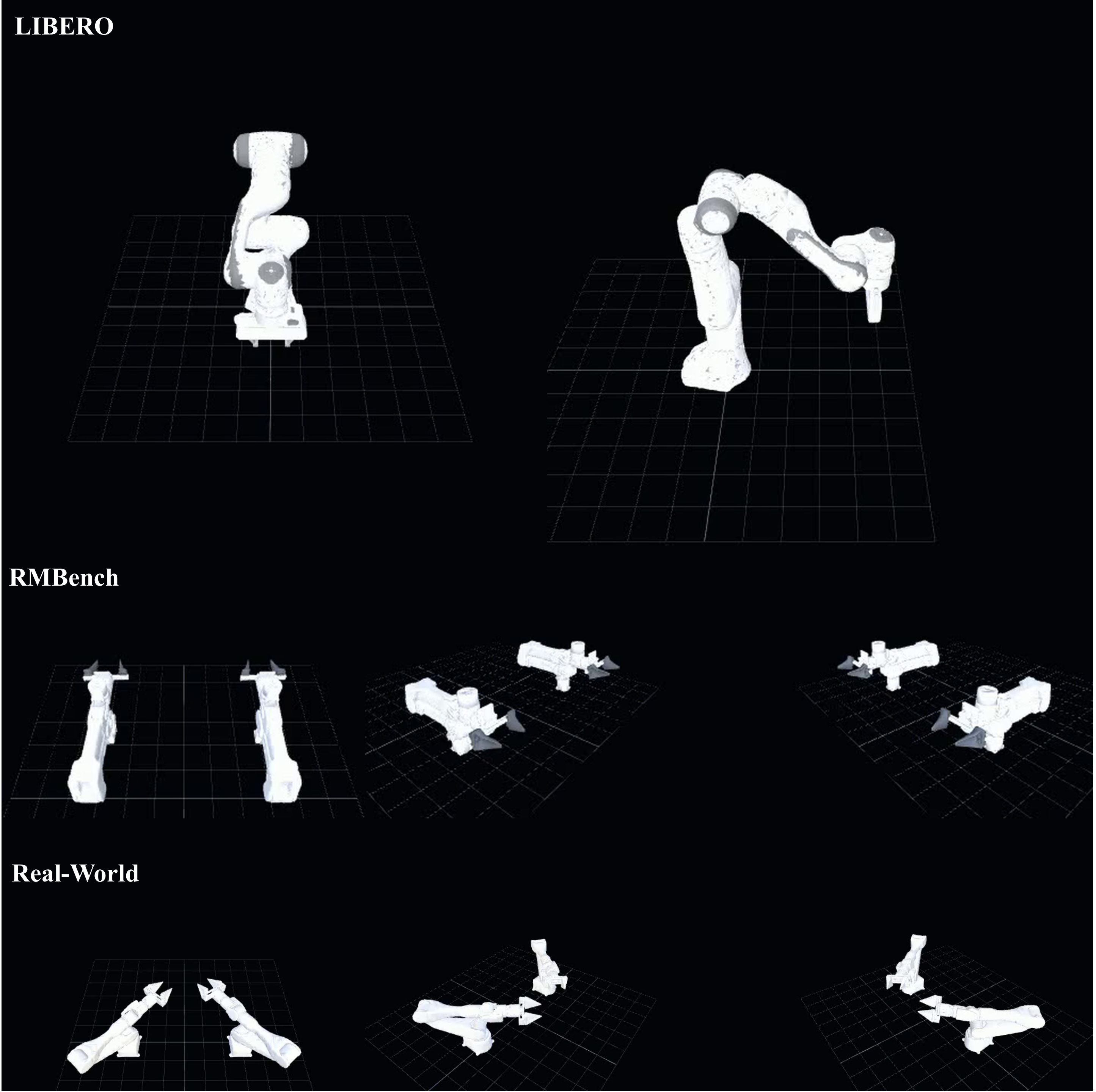}

\caption{\textbf{Visualization of visual-action frames.}
From top to bottom, the rows show examples from LIBERO, RMBench, and the real-world tasks.
The LIBERO example is rendered from two orthogonal viewpoints, and the other two examples are rendered from three mutually orthogonal viewpoints each.}

\label{fig:sup_va}
\end{figure}
\subsection{Visual-Action Construction}
As shown in Fig.~\ref{fig:sup_va}, we render visual-action frames from the robot states. We use two orthogonal views for LIBERO and three mutually orthogonal views for RMBench and the real-world tasks. In all three settings, we choose the primary rendering view to roughly match the viewing direction of one camera in the original videos. This correspondence is limited to the general viewing direction, with no precise alignment between the rendered and recorded views. The remaining orthogonal views serve as complementary perspectives on the same robot state.

\begin{figure}[t]
\centering
\includegraphics[width=1.0\linewidth]{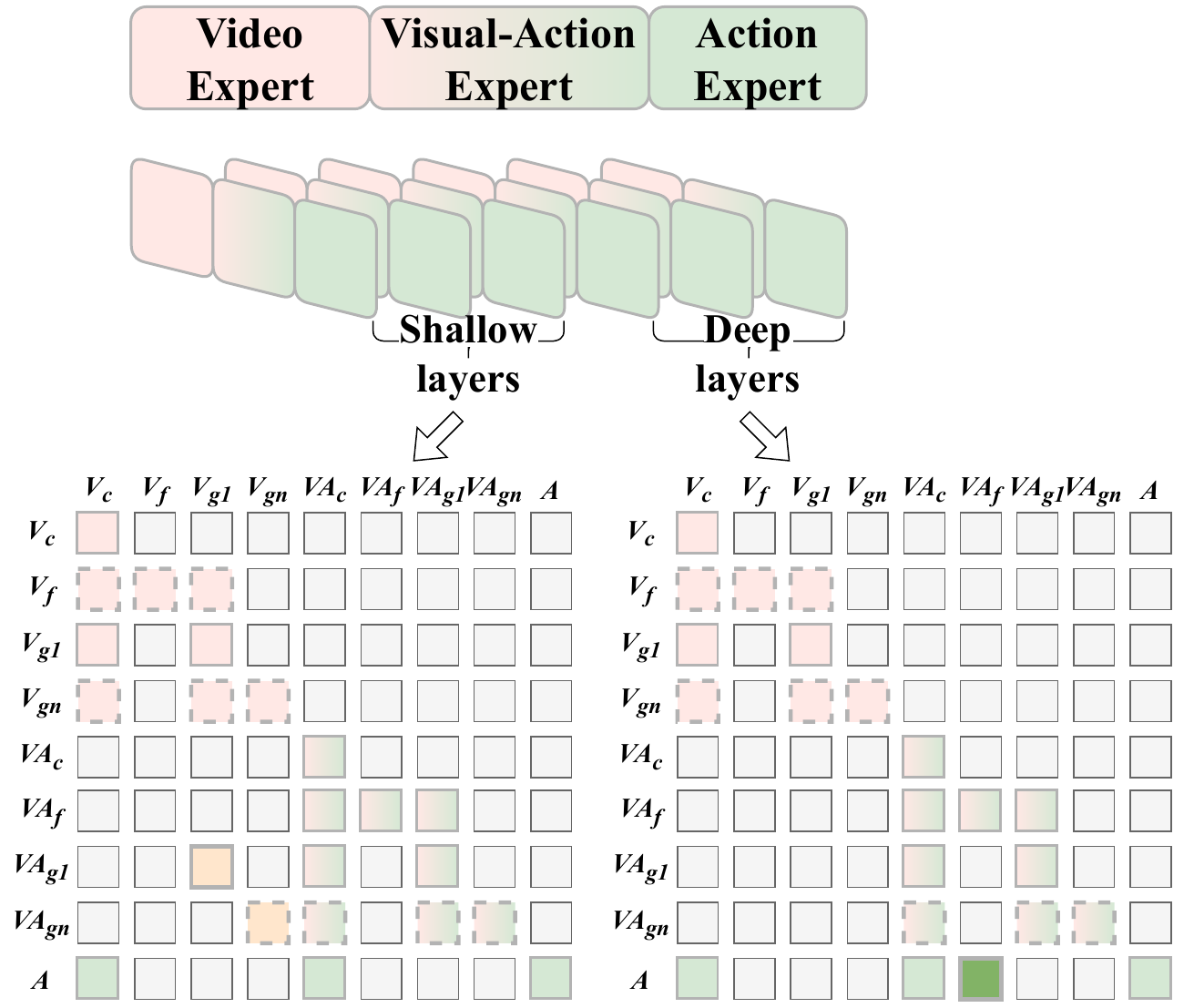}
\caption{\textbf{Layer-Grouped Attention Mask.} We split the network into two equal groups and apply a distinct attention mask to each group. The mask on the left is used for the first 15 (shallow) layers, and the mask on the right is used for the last 15 (deep) layers. Each group shares a fixed mask across its layers, with the mask switching between layers 15 and 16.}

\label{fig:sup_gm}
\end{figure}

\section{Model Details}
\label{sec:sup_method}
We initialize the video expert from Wan2.2-5B and use a variational autoencoder (VAE) to encode and decode visual and visual-action frames. The video expert has 30 Transformer layers and a hidden dimension of 3,072. The visual-action and action experts follow the same architecture with a hidden dimension of 1,024. We initialize both experts by interpolating the Wan2.2-5B weights to match the smaller hidden dimension. The model jointly predicts four visual subgoals and their four corresponding visual-action subgoals. To improve training efficiency, we halve the spatial resolution of the visual subgoal frames and all visual-action frames. As shown in Fig.~\ref{fig:sup_gm}, we use two cross-modal attention masks across the 30 Transformer layers. The first 15 layers share the shallow-layer mask (left), and the last 15 layers share the deep-layer mask (right).

\section{Training \& Evaluation Details}
\label{sec:sup_train_inf}

\subsection{Training hyperparameters}
\begin{table}[t]
  \centering
  \footnotesize
  \caption{\textbf{Training Hyperparameters for LIBERO.} Hyperparameters followed by (S2) are specific to Stage 2. All other settings are shared across both stages.}
  \label{tab:libero_training_hy}
  \setlength{\tabcolsep}{4pt}
  \begin{tabularx}{\linewidth}{Xc}
    \toprule
    Hyperparameter & Value \\
    \midrule
    Training epochs per stage & 10 \\
    Learning rate for the video and visual-action experts & $5 \times 10^{-5}$ \\
    Learning rate for the action expert learning rate & $1 \times 10^{-4}$ (S2) \\
    Learning rate schedule & Cosine \\
    Global batch size & 128 \\
    Weight decay & 0.01 \\
    Maximum gradient norm & 1.0 \\
    Loss weights (video / visual-action / action) & 1.0 / 1.0 / 1.0 (S2) \\
    Number of predicted visual subgoals & 4 \\
    Number of views (video / visual-action) & 2 / 2 \\
    Original video frame resolution & $256 \times 512$ \\
    Subgoal and visual-action frame resolution & $128 \times 256$ \\
    \bottomrule
  \end{tabularx}
\end{table}
\noindent\textbf{LIBERO.} On LIBERO, we jointly train SVA-WAM on all tasks from the four suites using a two-stage training procedure. As shown in Tab.~\ref{tab:libero_training_hy}, each stage runs for 10 epochs with a global batch size of 128. The model predicts four long-horizon visual subgoals and an action chunk of 32 time steps. The original video frames have a resolution of \(256 \times 512\) after concatenating the two camera views. To improve training efficiency, we use a resolution of \(128 \times 256\) for both the ground-truth subgoal and visual-action frames. Each visual-action frame combines two orthogonal views. The primary rendering view only approximately matches the viewing direction of one original camera.

\begin{table}[t]
  \centering
  \footnotesize
  \caption{\textbf{Training Hyperparameters for RMBench.}}
  \label{tab:rmbench_training_hy}
  \setlength{\tabcolsep}{4pt}
  \begin{tabularx}{\linewidth}{Xc}
    \toprule
    Hyperparameter & Value \\
    \midrule
    Training steps per task & 1,500 \\
    Learning rate & $1 \times 10^{-4}$ \\
    Learning rate schedule & Cosine \\
    Global batch size & 128 \\
    Weight decay & 0.01 \\
    Maximum gradient norm & 1.0 \\
    Loss weights (video / visual-action / action) & 1.0 / 1.0 / 1.0 \\
    Number of predicted visual subgoals & 4 \\
    Number of views (video / visual-action) & 3 / 3 \\
    Original video frame resolution & $384 \times 320$ \\
    Subgoal and visual-action frame resolution & $192 \times 160$ \\
    \bottomrule
  \end{tabularx}
\end{table}
\noindent\textbf{RMBench.} On RMBench, we train a separate SVA-WAM model for each task in a single stage, as shown in Tab.~\ref{tab:rmbench_training_hy}. Each model is trained for 1,500 steps with a global batch size of 128 and predicts four visual subgoals and an action chunk of 32 time steps. The original video frames have a resolution of \(384 \times 320\) after concatenating the three camera views. To improve training efficiency, we halve both spatial dimensions for the ground-truth subgoal and visual-action frames, giving both a resolution of \(192 \times 160\) after view concatenation. Each visual-action frame combines three mutually orthogonal views.

\begin{table}[t]
  \centering
  \footnotesize
  \caption{\textbf{Training Hyperparameters for Real-World Tasks}. Hyperparameters followed by (S2) are specific to Stage 2. All other settings are shared across both stages.}
  \label{tab:real_training_hy}
  \setlength{\tabcolsep}{4pt}
  \begin{tabularx}{\linewidth}{Xc}
    \toprule
    Hyperparameter & Value \\
    \midrule
    Training epochs per stage & 10 \\
    Video and visual-action expert learning rate & $5 \times 10^{-5}$ \\
    Action expert learning rate & $1 \times 10^{-4}$ (S2) \\
    Learning rate schedule & Cosine \\
    Global batch size & 256 \\
    Weight decay & 0.01 \\
    Maximum gradient norm & 1.0 \\
    Loss weights (video / visual-action / action) & 1.0 / 1.0 / 1.0 (S2) \\
    Number of predicted visual subgoals & 4 \\
    Number of views (video / visual-action) & 3 / 3 \\
    Original video frame resolution & $384 \times 320$ \\
    Subgoal and visual-action frame resolution & $192 \times 160$  \\
    \bottomrule
  \end{tabularx}
\end{table}
\noindent\textbf{Real-World Tasks.} In our real-world experiments, we collect 240, 210, and 264 demonstrations for Place Cup, Build Block, and Tidy Tableware, respectively. We train a separate SVA-WAM model for each task in two stages, with 10 epochs per stage and a global batch size of 256. Each model predicts four visual subgoals and an action chunk of 32 time steps. The original video frames have a resolution of \(384 \times 320\) after concatenating the three camera views. Each visual-action frame also combines three views rendered along mutually orthogonal directions. Tab.~\ref{tab:real_training_hy} lists the training hyperparameters. For the baselines, we train \(\pi_{0.5}\) for 30$k$ steps with a global batch size of 128 and train Fast-WAM for 20 epochs, using the same values as SVA-WAM for the hyperparameters shared by both methods.

\subsection{Evaluation Settings}
\noindent\textbf{LIBERO and LIBERO-Plus.} On LIBERO, we run 50 rollouts for each of the 40 tasks across the four suites, for a total of 2,000 rollouts. We use the same image resolutions as during training. At each inference call, the model uses 10 denoising steps to generate a 32-step action chunk. We execute the first 10 steps of the chunk before replanning. On LIBERO-Plus, we evaluate SVA-WAM zero-shot using the checkpoint trained on LIBERO. Following the official evaluation protocol, we run one rollout per test case, for a total of 10,030 rollouts. The model again uses 10 denoising steps per inference call, and we execute the first 5 steps of each predicted action chunk before replanning. For both benchmarks, we report the success rate as the percentage of rollouts that complete the task successfully.

\noindent\textbf{RMBench.} On RMBench, we follow the official evaluation protocol and run 100 rollouts per task. We compute the percentage of successful rollouts for each task and report the average across tasks. The model uses 10 denoising steps per inference call, and we execute the first 24 steps of each predicted 32-step action chunk before replanning.

\begin{figure}[t]
\centering
\includegraphics[width=1.0\linewidth]{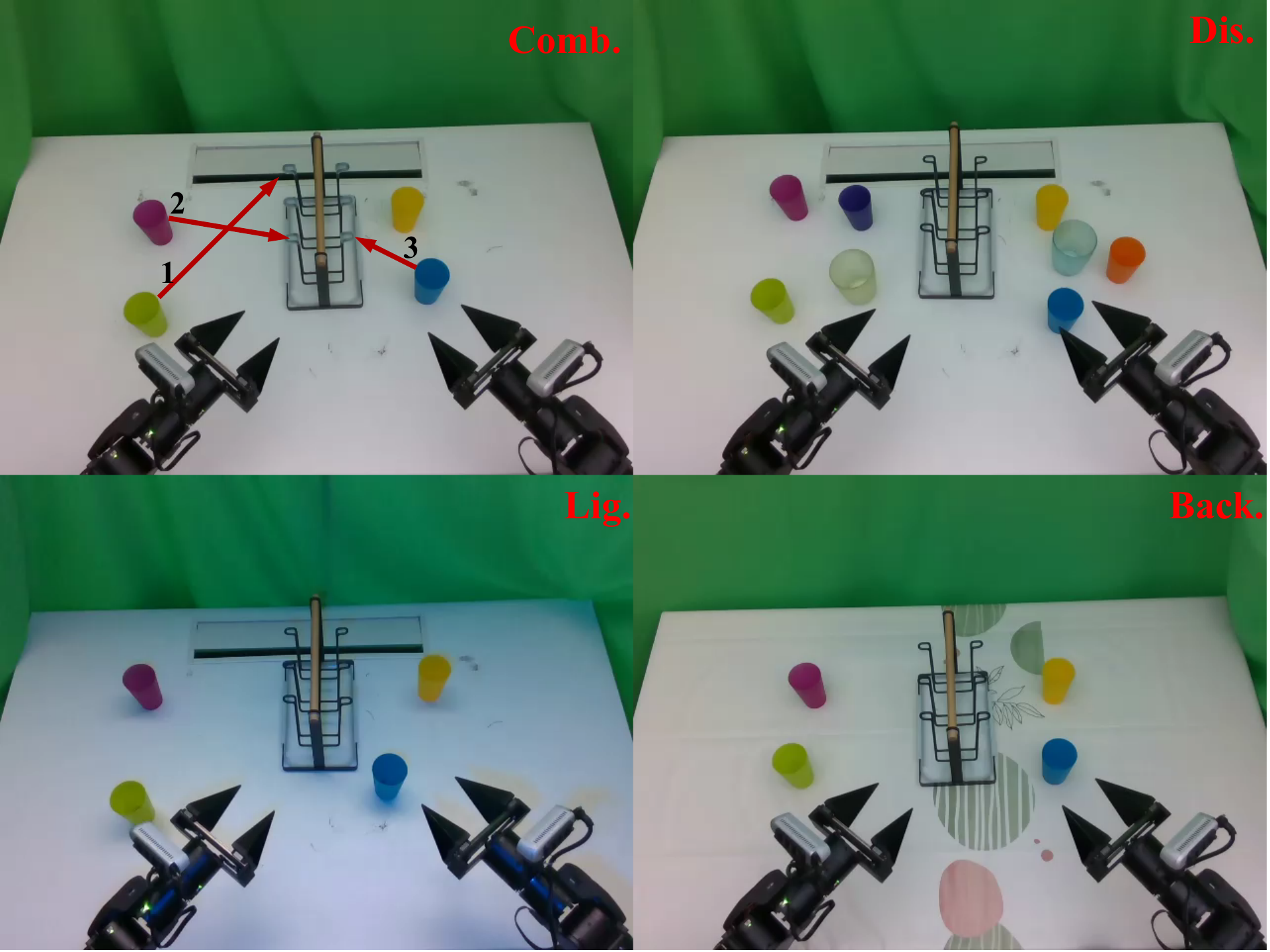}
\caption{\textbf{Out-of-distribution evaluation setups for Place Cup.}
The panels show an unseen combination of cup placement order and target positions, distractor objects absent from the demonstrations, altered lighting, and a different tabletop background.}

\label{fig:sup_ood}
\end{figure}

\noindent\textbf{Real-World Tasks.} For each of the three real-world tasks, we run 12 trials for SVA-WAM and each baseline. We report two metrics: success rate and progress score. Success rate is the percentage of trials in which the robot completes the entire task. The progress score measures partial completion and provides a finer-grained measure of performance. For Place Cup, the robot must place three cups in a prescribed order. Each cup placed in the correct order earns 10 points, for a maximum of 30 points. For Tidy Tableware, each correctly placed item earns 10 points, while each grasping error incurs a 2-point penalty and each placement error incurs a 5-point penalty. To evaluate generalization, we use four out-of-distribution (OOD) conditions: unseen combinations of placement orders and positions, distractors absent from the demonstrations, altered lighting, and a different tabletop background. The setups are shown in Fig.~\ref{fig:sup_ood}. We evaluate the OOD conditions using the same protocol as in the standard setting. At each inference call, SVA-WAM uses 10 denoising steps and executes the first 20 steps of the predicted action chunk before replanning.

\begin{table}[t]
\centering
\footnotesize
\caption{\textbf{Inference cost and task success rates.} We measure inference latency (ms) and computational cost (TFLOPs) per action chunk on LIBERO for Fast-WAM and SVA-WAM. We report task success rates (\%) on LIBERO and LIBERO-Plus.}
\label{tab:inference_comparison}

\setlength{\tabcolsep}{5pt}
\begin{tabularx}{\linewidth}{l|cccc}
\toprule
\textbf{Method}
& \textbf{Latency (ms)}
& \textbf{TFLOPs}
& \textbf{LIBERO}
& \textbf{LIBERO-Plus} \\
\midrule
Fast-WAM
& \textbf{356.01} & \textbf{2.12}  & 97.6 & 53.8 \\
\textbf{SVA-WAM}
& 423.75 & 3.28 & \textbf{98.8} & \textbf{74.8} \\
\bottomrule
\end{tabularx}
\end{table}
\section{Model Efficiency}
We measure per-action-chunk inference latency and computational cost for SVA-WAM and Fast-WAM on LIBERO using an NVIDIA PRO 6000 GPU. As shown in Tab.~\ref{tab:inference_comparison}, Fast-WAM takes 356.01~ms and requires \(2.12 \times 10^{12}\) FLOPs per action chunk. SVA-WAM takes 423.75~ms and requires \(3.28 \times 10^{12}\) FLOPs per action chunk. This additional computation accompanies substantial performance gains, particularly in zero-shot generalization on LIBERO-Plus.

\section{Additional Visualizations}
\begin{figure*}[t]
\centering
\includegraphics[width=1.0\linewidth]{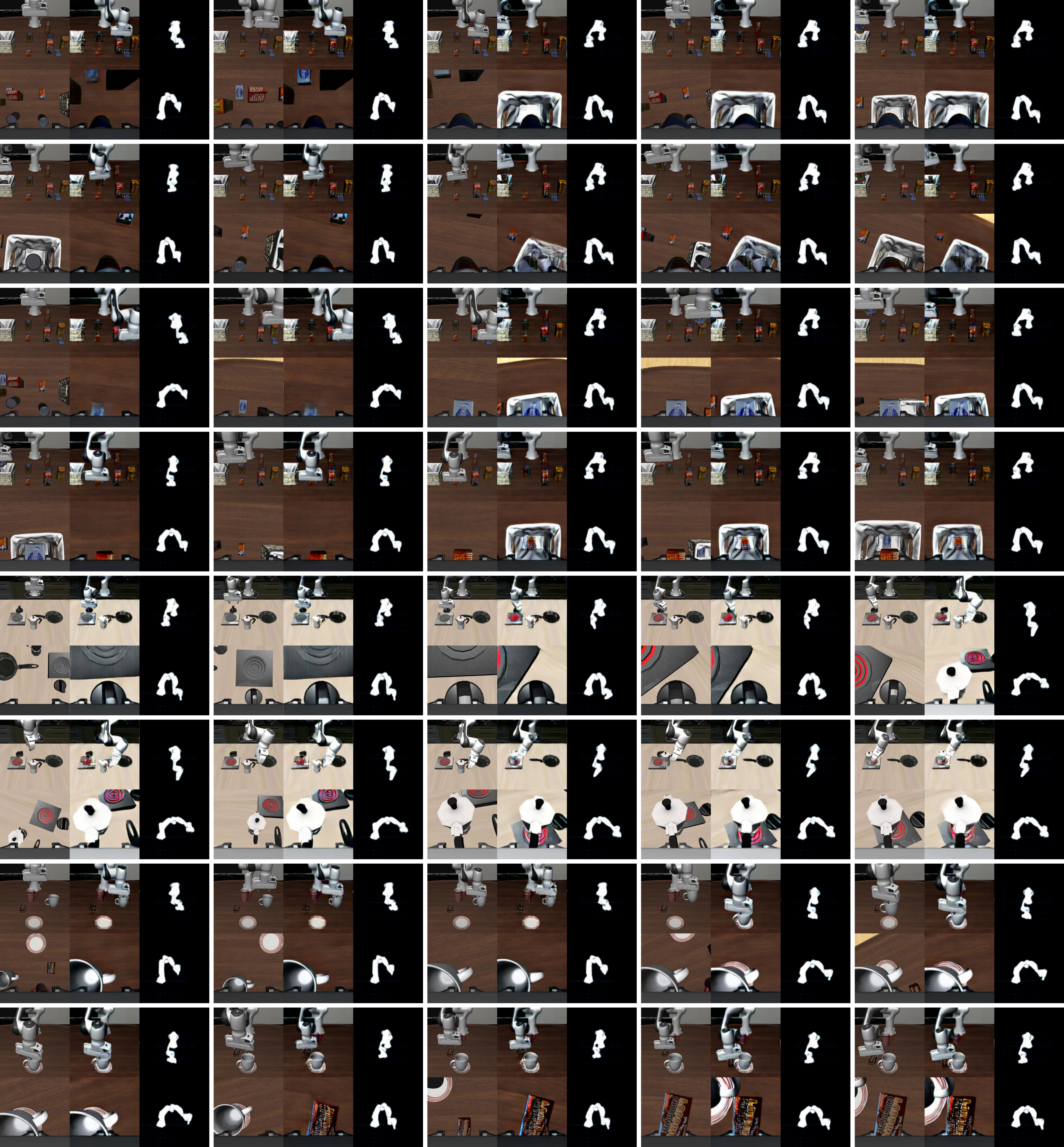}
\caption{\textbf{First-subgoal predictions on LIBERO.}
Each row shows successive prediction points in a rollout, ordered from left to right. At each point, we show the current observation, the predicted first visual subgoal, and the corresponding predicted visual-action subgoal.}

\label{fig:sup_lg}
\end{figure*}

\begin{figure*}[t]
\centering
\includegraphics[width=1.0\linewidth]{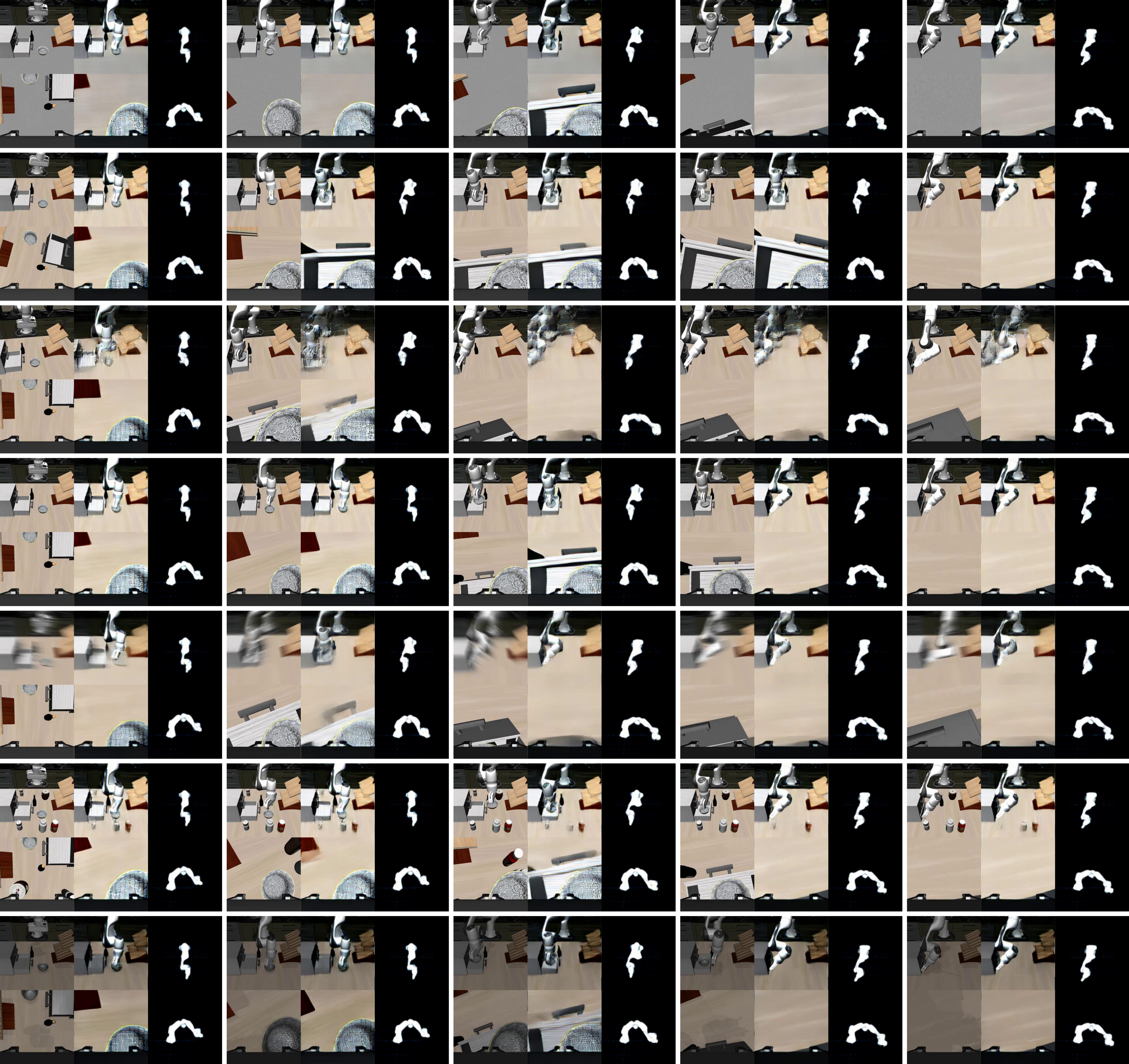}
\caption{\textbf{Zero-shot first-subgoal predictions on LIBERO-Plus.}
Each row shows successive prediction points in a rollout using the checkpoint trained on LIBERO. At each point, we show the current observation, the predicted first visual subgoal, and the corresponding predicted visual-action subgoal.}

\label{fig:sup_lpg}
\end{figure*}

\begin{figure*}[t]
\centering
\includegraphics[width=1.0\linewidth]{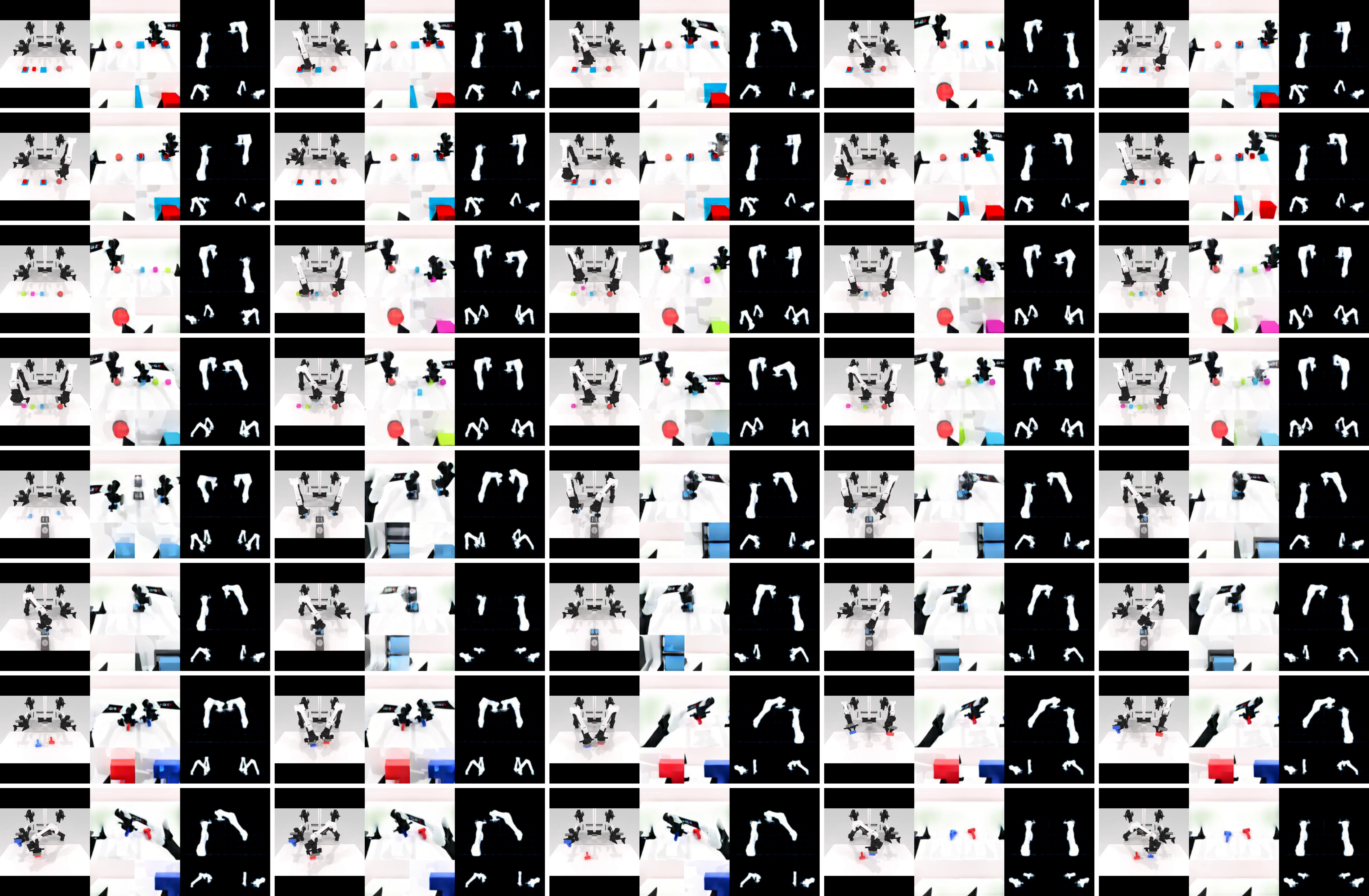}
\caption{\textbf{First-subgoal predictions on RMBench.}
Each row shows successive prediction points in a rollout, ordered from left to right.
At each point, we show the current observation, the predicted first visual subgoal, and the corresponding predicted visual-action subgoal.}

\label{fig:sup_rmg}
\end{figure*}

\begin{figure*}[t]
\centering
\includegraphics[width=1.0\linewidth]{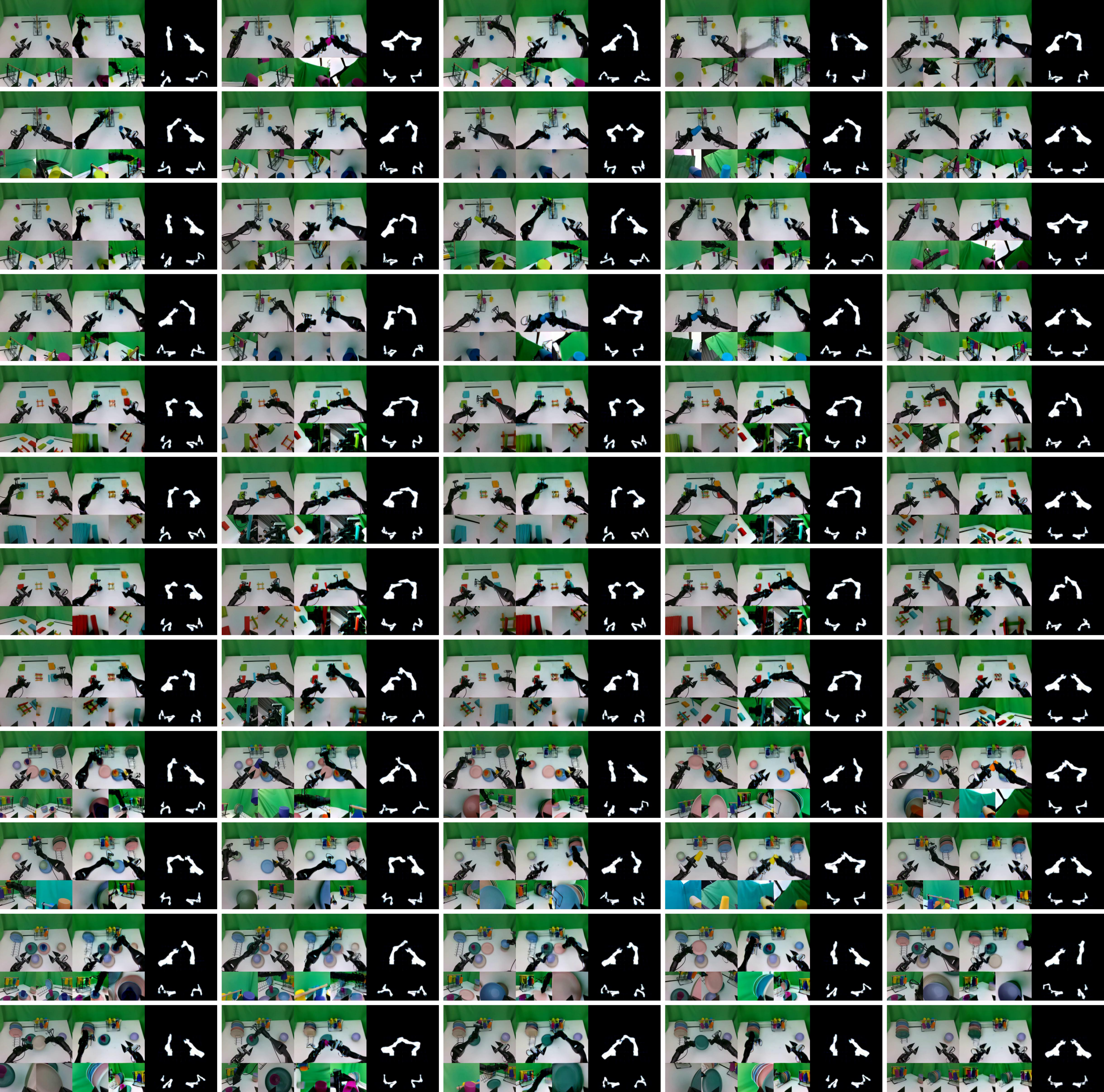}
\caption{\textbf{First-subgoal predictions on real-world tasks.}
Each row shows successive prediction points in a real-robot rollout, ordered from left to right.
At each point, we show the current observation, the predicted first visual subgoal, and the corresponding predicted visual-action subgoal.}

\label{fig:sup_rg}
\end{figure*}

Figs.~\ref{fig:sup_lg},~\ref{fig:sup_lpg},~\ref{fig:sup_rmg} and \ref{fig:sup_rg} show SVA-WAM's predictions of the first visual subgoal and its corresponding visual-action subgoal on LIBERO, LIBERO-Plus, RMBench, and the real-world tasks, respectively. In these examples, the predictions capture the intended next subgoal in both modalities, providing guidance for action generation.
\label{sec:sup_exp}

\end{document}